\documentclass[journal]{IEEEtran}

\IEEEoverridecommandlockouts                              %

\usepackage{tcolorbox}
\tcbuselibrary{theorems}
\newtcbtheorem{boxtheorem}{\it Theorem}{colback=gray!30, colframe=black!60, title=\textit}{thm}

\newcommand{\website}{https://xiaoyi-cai.github.io/evora}

\title{\LARGE \bf
EVORA: Deep Evidential Traversability Learning\\for Risk-Aware Off-Road Autonomy%
}
\author{Xiaoyi Cai$^1$, Siddharth Ancha$^1$, Lakshay Sharma$^1$, Philip R. Osteen$^{2}$,\\Bernadette Bucher$^3$, Stephen Phillips$^3$, Jiuguang Wang$^3$, Michael Everett$^4$, Nicholas Roy$^1$, and Jonathan P. How$^1$%
{
    \\
    \vspace*{1.2em}
    \small Project website: \href{\website}{\texttt{\color{deeppink}\website}}
    \\
    \vspace*{-2em}
}
\thanks{\edit{This research was sponsored by ARL grant W911NF-21-2-0150 and by ONR grant N00014-18-1-2832. The experiments using the Spot robot was completed during Xiaoyi Cai's internship at the Boston Dynamics AI Institute (BDAI). The authors thank Harrison Busa, Velin Dimitrov, Brennan Vanderlaan, and Leonor Fermoselle from BDAI for their help on the Spot robot's payload, and Brian Okorn and Surya Singh from BDAI for their feedback on the paper. The authors also thank Andrew Fishberg from MIT for his help on the RC car.}}
\thanks{$^1$Massachusetts Institute of Technology, Cambridge, MA 02139, USA. {\tt\small\{xyc, sancha, lakshays, nickroy, jhow\}@mit.edu}.}
\thanks{$^2$DEVCOM Army Research Laboratory, Adelphi, MD 20783, USA. {\tt\small philip.r.osteen.civ@army.mil}.}
\thanks{$^3$Boston Dynamics AI Institute, Cambridge, MA 02142, USA. {\tt\small \{bbucher, sphillips, jw\}@theaiinstitute.com}.}
\thanks{$^4$Northeastern University, Boston, MA 02115, USA. {\tt\small  m.everett@northeastern.edu}.}
}

\usepackage{amsmath,amsthm,amssymb, amsfonts} %
\usepackage{bbold}
\usepackage{mathtools}
\usepackage{graphicx}
\usepackage{xcolor}

\usepackage[export]{adjustbox} %

\usepackage{pdfpages}
\usepackage{bm}
\usepackage{url}
\usepackage{enumerate}
\usepackage{float,balance}
\usepackage[sort, compress]{cite}
\usepackage{cases,balance}
\usepackage[pdftex, pdfstartview={FitV}, pdfpagelayout={TwoColumnLeft},bookmarksopen=true,plainpages = false, colorlinks=true, linkcolor=black, citecolor = black, urlcolor = black,filecolor=black , pagebackref=false,hypertexnames=false, plainpages=false, pdfpagelabels ]{hyperref}

\usepackage[capitalise]{cleveref}

\usepackage[font=footnotesize]{caption}

\usepackage{array}
\usepackage{booktabs}
\usepackage{colortbl}
\usepackage{multirow} %
\usepackage{threeparttable} %

\usepackage{placeins}
\usepackage{ulem}  %

\usepackage{comment} %
\excludecomment{comments} %

\usepackage{algorithm}
\usepackage[noend]{algpseudocode} %
\definecolor{commentclr}{RGB}{110, 149, 204}
\definecolor{deeppink}{rgb}{1.0, 0.08, 0.58}

\makeatletter
\newcommand\fs@spaceruled{\def\@fs@cfont{\bfseries}\let\@fs@capt\floatc@ruled
  \def\@fs@pre{\vspace{0.6\baselineskip}\hrule height.8pt depth0pt \kern2pt}%
  \def\@fs@post{\kern2pt\hrule\relax}%
  \def\@fs@mid{\kern2pt\hrule\kern2pt}%
  \let\@fs@iftopcapt\iftrue}
\makeatother

\usepackage{tikz}

\usepackage{tcolorbox}
\newtcbox{\dashedbox}[1][]{
  math upper,
  baseline=0.4\baselineskip,
  nobeforeafter,
  colback=white,
  boxrule=0pt,
  enhanced jigsaw,
  boxsep=0pt,
  top=2pt,
  bottom=2pt,
  left=2pt,
  right=2pt,
  borderline horizontal={0.5pt}{0pt}{dashed},
  borderline vertical={0.5pt}{0pt}{dashed},
  #1
}

\newcommand{\setO}{\boldsymbol{O}} %

\newcommand{\Param}{\boldsymbol{\psi}}
\newcommand{\setParam}{\boldsymbol{\Psi}}

\newcommand{\setX}{\mathbf{X}} %

\newcommand{\done}[1]{\mathbb{1}^{\text{done}}(#1)} %

\newcommand{\lcvar}[2]{{\text{CVaR}_{#1}^{\leftarrow}(#2)}}
\newcommand{\rcvar}[2]{{\text{CVaR}_{#1}^{\rightarrow}(#2)}}
\newcommand{\lvar}[2]{{\text{VaR}_{#1}^{\leftarrow}(#2)}}
\newcommand{\rvar}[2]{{\text{VaR}_{#1}^{\rightarrow}(#2)}}

\newcommand{\cvarcost}{{CVaR-Cost}}
\newcommand{\cvardyn}{{CVaR-Dyn}}

\newcommand{\digammafun}{{\mathbb{\Psi}}}

\newcommand{\ce}{{\text{CE}}}
\newcommand{\uce}{{\text{UCE}}}
\newcommand{\emd}{{\text{EMD}}}
\newcommand{\emdsq}{{\text{EMD}^2}}
\newcommand{\uemdsq}{{\text{UEMD}^2}}

\newcommand{\dir}{{\text{Dir}}}
\newcommand{\cat}{{\text{Cat}}}

\newcommand{\covariance}{{\text{Cov}}}
\newcommand{\variance}{{\text{Var}}}
\newcommand{\cumsum}{{\text{cs}}}

\newcommand{\E}{{\mathbb{E}}}

\newcommand{\R}{\mathbb{R}}

\newcommand{\red}[1]{{\color{red}#1}}

\newcommand{\edit}[1]{\textcolor{black}{#1}}

\newcommand{\tr}{^\top}
\newcommand*{\defeq}{:=}

\theoremstyle{plain}%

\newtheorem{theorem}{Theorem}

\theoremstyle{remark}

\theoremstyle{definition}

\graphicspath{{Figs/}}

\begin{document}
\bstctlcite{IEEEexample:BSTcontrol} %
\maketitle
\thispagestyle{empty}
\pagestyle{empty}

\begin{abstract}
Traversing terrain with good traction is crucial for achieving fast off-road navigation. Instead of manually designing costs based on terrain features, existing methods learn terrain properties directly from data via self-supervision to automatically penalize trajectories moving through undesirable terrain, but challenges remain to properly quantify and mitigate the risk due to uncertainty in learned models. To this end, this work proposes a unified framework to learn uncertainty-aware traction model and plan risk-aware trajectories.
For uncertainty quantification, we efficiently model both aleatoric and epistemic \edit{uncertainty} by learning discrete traction distributions and probability densities of the traction predictor's latent features. Leveraging evidential deep learning, we parameterize Dirichlet distributions with the network outputs and propose a novel uncertainty-aware squared Earth Mover's distance loss with a closed-form expression that improves learning accuracy and navigation performance.
\edit{For risk-aware navigation, the proposed planner simulates} state trajectories with the worst-case expected traction to handle aleatoric uncertainty, and penalizes trajectories moving through terrain with high epistemic uncertainty. Our approach is extensively validated in simulation and on wheeled and quadruped robots, showing improved navigation performance compared to methods that assume no slip, assume the expected traction, or optimize for the worst-case expected cost.
\end{abstract}

\section{Introduction}

\begin{figure}[t]
\centering
\includegraphics[
width=\linewidth, trim={0cm 0cm 0cm 0cm},clip]{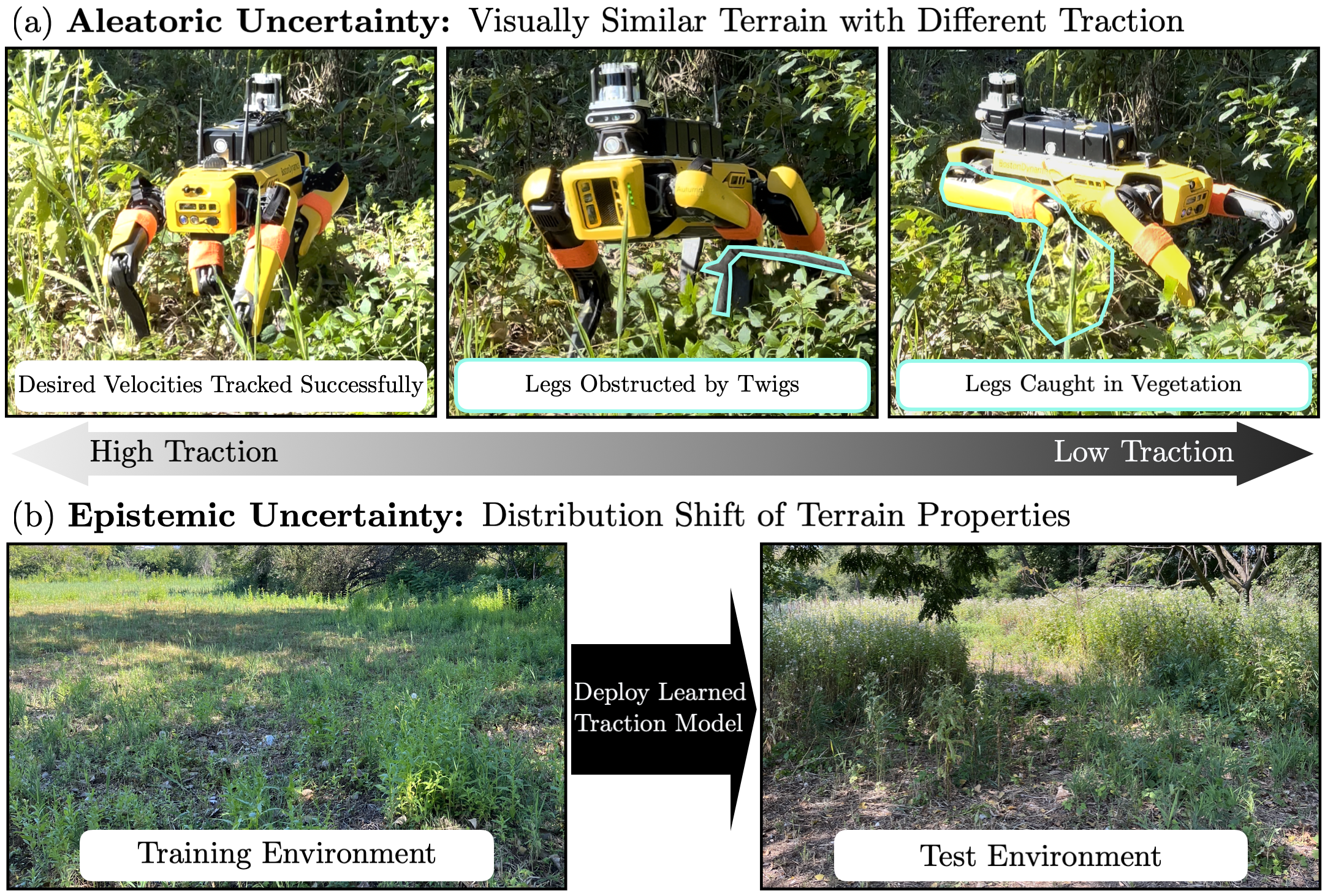}
\caption{
This work proposes to learn terrain \textit{traction}, the ratio between achieved and commanded velocities, while quantifying the uncertainty in the learned model to plan risk-aware trajectories. (a)~\textit{Aleatoric uncertainty} is the inherent and irreducible uncertainty due to partial observability. For example, visually similar terrain may have different traction values due to complex interactions between the robot and vegetation. (b)~\textit{Epistemic uncertainty} is the model uncertainty due to distribution shift between training and test environments, limiting the reliability of the learned model at test time.
}
\label{fig:motivation_uncertainty}
\vspace*{-0.2in}
\end{figure}

Autonomous robots are increasingly being deployed in harsh off-road environments like mines, forests, and deserts~\cite{fan2021step, ruetz2023foresttrav, meng2023terrainnet}, where both geometric and semantic understanding of the environments is required to identify non-geometric hazards (e.g., mud puddles, slippery surfaces) and geometric non-hazards (e.g., tall grass and foliage) \edit{in order to achieve reliable navigation}. To this end, recent approaches manually assign navigation costs based on semantic classification of the terrain~\cite{shaban2022semantic, chen2022cali, guan2022ga}, requiring significant human expertise to label and train a classifier sufficiently accurate and rich in order to achieve desired risk-aware behaviors.
Alternatively, self-supervised learning can be used to learn a model of traversability directly from navigation data~\cite{kahn2021badgr, yao2022rca, zurn2020self} \edit{to automatically assign higher costs for undesirable terrain during planning.}
\edit{Because self-supervised data collection in the real world can be slow and expensive, collecting more data is not beneficial unless we properly quantify and mitigate the risk due to uncertainty in the learned models. Uncertainty manifests in two forms as illustrated in Fig.~\ref{fig:motivation_uncertainty} in the context of off-road navigation.}
\textit{Aleatoric uncertainty} is the inherent and irreducible uncertainty due to partial observability. For example, two patches of terrain may be indistinguishable to the onboard sensors but lead to different vehicle behaviors---such uncertainty cannot be reduced by collecting more data.
\textit{Epistemic uncertainty} is due to out-of-distribution (OOD) inputs encountered at test time that are not well-represented in the training data.
\edit{Because it is often undesirable to collect OOD data in dangerous situations such as collisions and falling at the edge of a cliff, there can exist a large gap between training datasets and the various real-world scenarios encountered by the robot.}
Most existing work in off-road navigation has focused on either aleatoric uncertainty~\cite{ewen2022probfriction, cai2022risk} \edit{by learning distributions of system parameters instead of point estimates}, or epistemic uncertainty~\cite{Frey-RSS-23, schmid2022self, lee2023learning, endo2023risk} \edit{by identifying OOD terrain}, but limited effort has been made to \edit{quantify} both types of \edit{uncertainty and mitigate the associated risk during planning}.

\begin{figure*}[t]
\centering
  \includegraphics[width=0.9\textwidth]{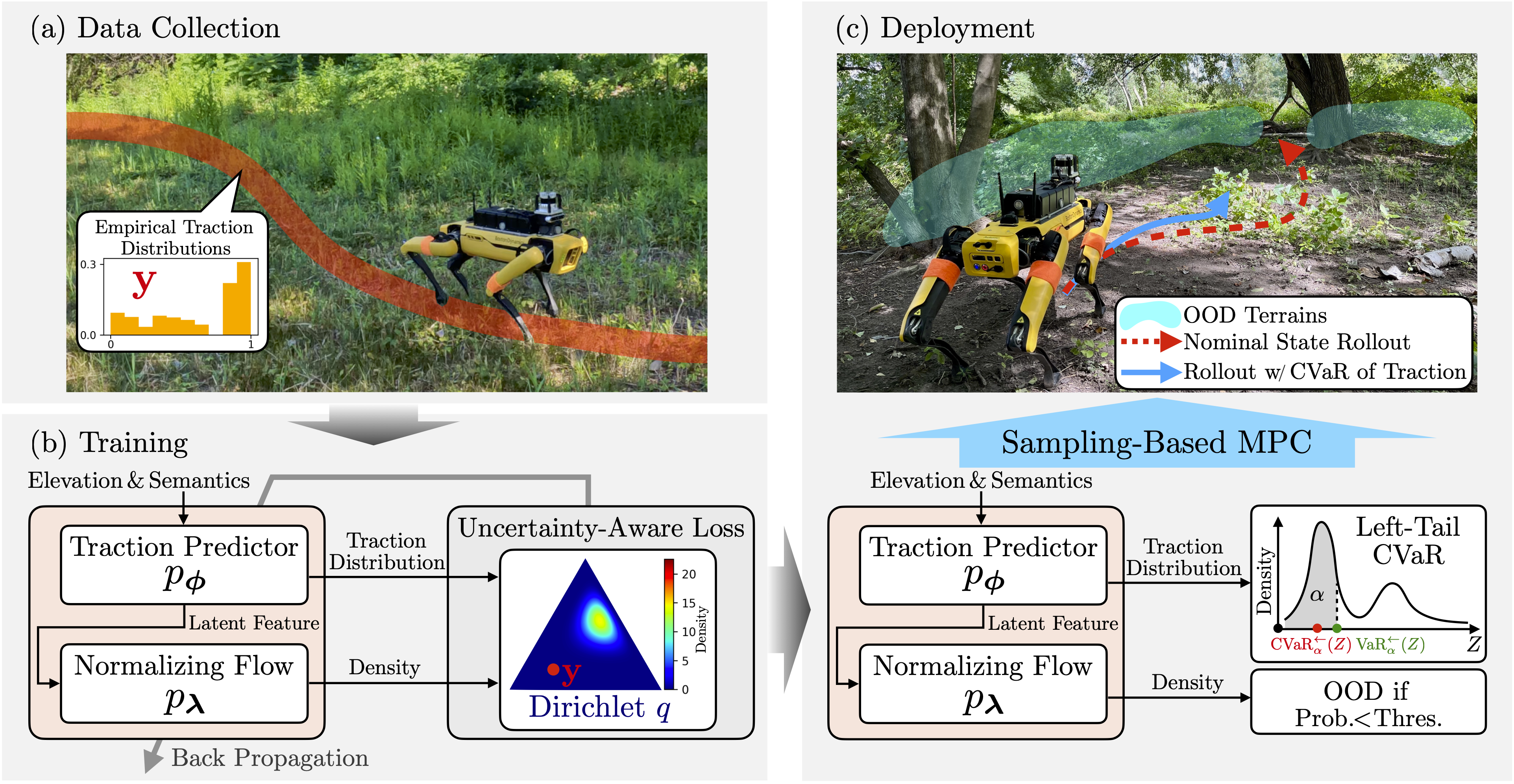}
  \captionof{figure}{
Overview of the proposed \textit{uncertainty-aware traversability learning} and \textit{risk-aware navigation} methods. 
\edit{
(a)~For data collection, we drive the robot over interesting terrain to record traction values, robot positions, and build a semantic elevation map. We generate training dataset offline by extracting semantic and elevation features of the terrain and estimating empirical traction distributions along the traversed path. 
}
(b)~Leveraging evidential deep learning~\cite{natpn}, we learn categorical distributions over discretized traction values to capture \textit{aleatoric uncertainty} and estimate \textit{epistemic uncertainty} by using a normalizing flow network~\cite{Kobyzev2021flow} to learn the densities of the traction predictor's latent features. 
The overall architecture is trained with the proposed uncertainty-aware loss defined for the Dirichlet distribution parameterized by the network outputs.
(c)~To handle aleatoric uncertainty, we propose a risk-aware planner that uses the left-tail conditional value at risk (CVaR) of the traction distribution to forward simulate the robot states when using the sampling-based model predictive control (MPC) method~\cite{williams2017information}. 
To handle epistemic uncertainty, we threshold the densities of the traction predictor's latent features in order to identify and avoid out-of-distribution (OOD) terrain with unreliable traction predictions via auxiliary planning costs.
} \label{fig:overall_approach}
\vspace*{-0.2in}
\end{figure*}

\edit{
To achieve fast and reliable off-road navigation, this work considers both the upstream \textit{uncertainty-aware traversability learning} problem and the downstream \textit{risk-aware navigation} problem. Recognizing the inter-dependence of the two problems, our proposed pipeline, \textbf{EVORA} (\textbf{EV}idential \textbf{O}ff-\textbf{R}oad \textbf{A}utonomy), tightly integrates the proposed uncertainty-aware traversability model into the the proposed risk-aware planner (see Fig.~\ref{fig:overall_approach} for an overview).
}
To plan fast trajectories, we model traversability with terrain \textit{traction} that captures the ``slip'' or the ratio between \edit{achieved and commanded} velocities (for example, wet terrain that causes the robot's wheels to slip and reduce its intended velocity has low traction). Moreover, we efficiently quantify both aleatoric and epistemic \edit{uncertainty} by learning the empirical traction distributions and probability densities of the traction predictor's latent features. 
\edit{
Because real world traction distributions may be multi-modal, as shown in {Fig.~\ref{fig:motivation_uncertainty}(a)} where vegetation with similar appearance may lead to different traction values, we learn categorical distributions over discretized traction values to capture multi-modality. 
}
By leveraging the evidential deep learning technique proposed in~\cite{natpn}, we parametrize Dirichlet distributions (the conjugate priors for the categorical distributions) with neural network (NN) outputs, and propose a novel uncertainty-aware loss based on the squared Earth Mover's distance~\cite{hou2016emd2}. Our loss, which can be computed efficiently in closed-form, better captures the relationship among discretized traction values than the conventional cross entropy-based losses~\cite{murphy2012machine}. 
To handle aleatoric uncertainty, we propose a risk-aware planner that simulates state trajectories using the worst-case expected traction, \edit{which is shown to achieve improved or competitive performance compared to} state-of-the-art methods that rely on the nominal traction~\cite{cai2022risk}, the expected traction~\cite{Gasparino2022wayfast} or that optimize for the worst-case expected cost~\cite{wang2021adaptive}. To mitigate the risk due to epistemic uncertainty, the proposed method imposes a confidence threshold on the densities of the traction predictor's latent space features to identify OOD terrain and avoid moving through it using auxiliary planning costs. The overall approach is extensively analyzed in simulation and hardware with wheeled and quadruped robots, demonstrating feasibility and improved navigation performance in practice.

\subsection{Related Work}
\subsubsection{\edit{Traversability Analysis}}
\edit{Suitability of terrain for navigation can be assessed} in various ways, e.g., based on proprioceptive measurements~\cite{oliveira2021three, otte2016recurrent}, geometric features~\cite{ruetz2023foresttrav, overbye2022g, fan2021step} and combinations of geometric and semantic features~\cite{jagan2023vern, shaban2022semantic, meng2023terrainnet} (see the survey in~\cite{papadakis2013terrain}). Due to the difficulty of hand-crafting planning costs based on  terrain features, self-supervised learning is increasingly being adopted to learn task-relevant traversability representations. 
For example, \edit{Li et al.~\cite{li2023seeing} proposed to learn the} support surfaces underneath dense vegetation for legged robot locomotion, and \edit{Gasparino et al.~\cite{Gasparino2022wayfast} modeled terrain traction that captures how well the robot can follow the desired velocities}. However, these methods do not account for the aleatoric and epistemic \edit{uncertainty} due to the noisiness and scarcity of real-world data.
To capture aleatoric uncertainty, \edit{Ewen et al.~\cite{ewen2022probfriction} and Cai et al.~\cite{cai2022risk} learned} multi-modal terrain properties via Gaussian mixture models or categorical distributions. 
To capture epistemic uncertainty, \edit{Frey et al.~\cite{Frey-RSS-23} and Schmid et al.~\cite{schmid2022self} measured} the trained NNs' ability to reconstruct terrain similar to the terrain types traversed in the past, and \edit{Seo et al.~\cite{seo2023scate} trained a binary classifier} for unfamiliar terrain.
In comparison, \edit{Endo et al.~\cite{endo2023risk} and Lee et al.~\cite{lee2023learning} leveraged} Gaussian Process (GP) regression \edit{to quantify epistemic uncertainty}, but they \edit{used} a homoscedastic noise model that assumes the noise variance is globally constant. While \edit{Murphy et al.~\cite{murphy2012creating} adopted} heteroscedastic GPs that can handle input-dependent noise, the predictive distributions are not analytically tractable and require approximations.
\edit{
In contrast, our work explicitly quantifies \edit{both the aleatoric and epistemic uncertainty} in the learned traction model that predicts the ratio between achieved and commanded velocities. While we learn traction just like Gasparino et al.~\cite{Gasparino2022wayfast}, our model is uncertainty-aware and can be used to achieve risk-aware navigation. 
In comparison, Frey et al.~\cite{Frey-RSS-23} used the difference between achieved and commanded velocities in the planning objective, but they assumed no slip when simulating the state rollouts. In contrast, our traction model can be used to simulate state rollouts under the worst-case expected traction condition, which is shown by our results to achieve better performance than methods that assume nominal traction when obtaining state rollouts.
}

\subsubsection{\edit{Uncertainty Quantification \& OOD Detection}}
Uncertainty quantification is well studied in the machine learning literature (see the survey in~\cite{gawlikowski2021survey}) with effective techniques such as Bayesian dropout~\cite{gal2016dropout}, model ensembles~\cite{osband2016deep}, and evidential methods~\cite{ulmer2023prior}.
In the off-road navigation literature, ensemble methods have been a popular choice~\cite{triest2023learning, lee2023AleatoricEpistemicKinodynamicModel, kim2023bridging}, because they typically outperform methods based on Bayesian dropout~\cite{lakshminarayanan2017simple}. In comparison, evidential methods are better suited for real-world deployment, because they only require a single network evaluation without imposing high computation or memory requirements. Therefore, we leverage the evidential method proposed \edit{by Charpentier et al.~\cite{natpn} to} directly parameterize the conjugate prior distribution of the target distribution with NN outputs \edit{in order to quantify both aleatoric and epistemic uncertainty.}
\edit{Moreover, we propose an uncertainty-aware loss based on the squared Earth Mover's Distance proposed by Hou et al.~\cite{hou2016emd2} to better capture the relationship among the discrete traction values,} resulting in more accurate traction predictions that in turn improve the downstream risk-aware planner's navigation performance.

\edit{
When deploying the learned traction model, we explicitly identify OOD terrain based on the estimated epistemic uncertainty, which is an instance of the general OOD detection problem (see the survey in~\cite{yang2021generalized}). For example, reconstruction-based method adopted by Seo et al.~\cite{seo2023learning} and density-based method adopted by Ancha et al.~\cite{ancha2024icra} have shown promising results for off-road navigation to identify unsafe terrain. Similar to Ancha et al.~\cite{ancha2024icra}, our approach is a density-based approach that explicitly captures the normalized probability density under the training data distribution. Alternatively, energy-based approaches proposed by Liu et al.~\cite{liu2020energy} and Grathwohl et al.~\cite{Grathwohl2020your} do not require explicit density normalization, and similar ideas have been adopted by Castaneda et al.~\cite{castaneda2023distribution} to avoid OOD states. 
Instead of solely focusing on OOD detection and mitigation, this work quantifies and mitigates the risk due to both aleatoric and epistemic uncertainty. 
While OOD terrain with high epistemic uncertainty should be avoided at test time, in-distribution terrain may still lead to high aleatoric uncertainty in the predicted traction due to complex vehicle-terrain interactions. Therefore, the risk due to aleatoric uncertainty should be mitigated separately to improve navigation performance by allowing the robot to trade off the likelihood of experiencing low traction with the potential time savings obtained from traversing terrain with uncertain traction.
}

\subsubsection{\edit{Risk-Aware Planning}}

\edit{The risk of traversing terrain with uncertain traversability values has been represented as costmaps by Fan et al.~\cite{fan2021learning} and Triest et al.~\cite{triest2023learning}}, where Conditional Value at Risk (CVaR) can be used to measure the cost of encountering worst-case expected failures, which satisfies a group of axioms important for rational risk assessment~\cite{majumdar2020should}. 
\edit{Instead of costmaps, navigation performance has also been assessed based on the expected future states by Gibson et al.~\cite{gibson2023multi} or the expected terrain traction by Gasparino et al.~\cite{Gasparino2022wayfast}.}
However, these methods rely on either the nominal or the expected system behavior, which may provide a poor indication of the actual performance when the vehicle-terrain interaction is noisy \edit{(i.e., high aleatoric uncertainty).}
\edit{Alternatively, Wang et al.~\cite{wang2021adaptive} proposed to directly optimize the CVaR of the planning objective, which can be estimated by evaluating each control sequence over samples of \edit{uncertain parameters}, but this approach is computationally expensive.}
\edit{Similar to our approach, the recent work of Lee et al.~\cite{lee2023AleatoricEpistemicKinodynamicModel} quantified both aleatoric and epistemic uncertainty using probabilistic ensembles~\cite{chua2018deep} and planned risk-aware trajectories by penalizing both types of \edit{uncertainty}, but it relied on the expected system behaviors. 
While we adopt a similar strategy used by Lee et al.~\cite{lee2023AleatoricEpistemicKinodynamicModel}  for handling epistemic uncertainty via auxiliary penalties, we use the worst-case expected system parameters for forward simulation to assess the risk due to aleatoric uncertainty. Our approach is computationally more efficient than the method proposed by Wang et al.~\cite{wang2021adaptive} and produces behaviors more robust to multi-modal terrain properties observed in the real world compared to methods proposed by Lee et al.~\cite{lee2023AleatoricEpistemicKinodynamicModel} and Gasparino et al.~\cite{Gasparino2022wayfast} that rely on the expected system behaviors.}

\subsection{Contributions}
This work proposes an off-road navigation pipeline that tightly integrates the solutions to the uncertainty-aware traversability learning problem and the risk-aware motion planning problem.
We explicitly quantify both the epistemic uncertainty to understand when the predicted traction values are unreliable due to novel terrain and the aleatoric uncertainty to enable the downstream planner to mitigate risk due to noisy traction estimates. 
The main contributions of this work are:
\begin{enumerate}
    \item A probabilistic traversability model based on traction distributions (aleatoric uncertainty), with the ability to identify unreliable predictions via the densities of the traction predictor's latent features (epistemic uncertainty).
    \item A novel uncertainty-aware loss based on the squared Earth Mover's Distance ($\emdsq$~\cite{hou2016emd2}) with a closed-form expression derived in this work that improves traction prediction accuracy, OOD detection performance, and downstream navigation performance when used together with the uncertainty-aware cross entropy loss~\cite{natpn}.
    \item A risk-aware planner based on the CVaR of traction to handle aleatoric uncertainty. Our planner outperforms methods that assume the nominal traction~\cite{cai2022risk} or the expected traction~\cite{Gasparino2022wayfast}, \edit{and achieves improved or competitive performance compared to the method optimizing for the CVaR of cost\cite{wang2021adaptive} in both} simulation and hardware.
    \item A further extension of the risk-aware planner to handle epistemic uncertainty by avoiding OOD terrain, \edit{which improves the navigation success rate in simulation and reduces human interventions in hardware experiments.}
\end{enumerate}

\noindent The preliminary conference version of this work appeared in~\cite{cai2022probabilistic}, \edit{which proposed to learn traction distributions and use CVaR of traction for planning.} This work extends the prior work by \edit{using the evidential learning technique proposed in~\cite{natpn} for model training, and deriving a new uncertainty-aware $\emdsq$ loss based on~\cite{hou2016emd2} to improve learning performance.} The new methods introduced in this work not only improve the accuracy of traction prediction and OOD detection but also leads to faster navigation. By adding extensive hardware experiments, this work provides stronger evidence of the performance improvements provided by the risk-aware planner proposed in the conference version~\cite{cai2022probabilistic} compared to the state-of-the-art methods~\cite{cai2022risk, Gasparino2022wayfast, wang2021adaptive}.

\section{Problem \edit{Overview}}
We consider the problem of motion planning for a ground vehicle whose dynamics depend on the terrain traction. Because traction values can be uncertain \edit{due to} rough terrain and imperfect sensing, we model traction values as random variables whose distributions can be estimated from sensor data using a learned model. \edit{Next, we introduce the dynamical models and the planning objective under consideration.} 

\subsection{\edit{Dynamical Models with Traction Parameters}}

\begin{figure}[t]
\centering
\includegraphics[width=\linewidth, trim={0cm 0cm 0cm 0cm},clip]{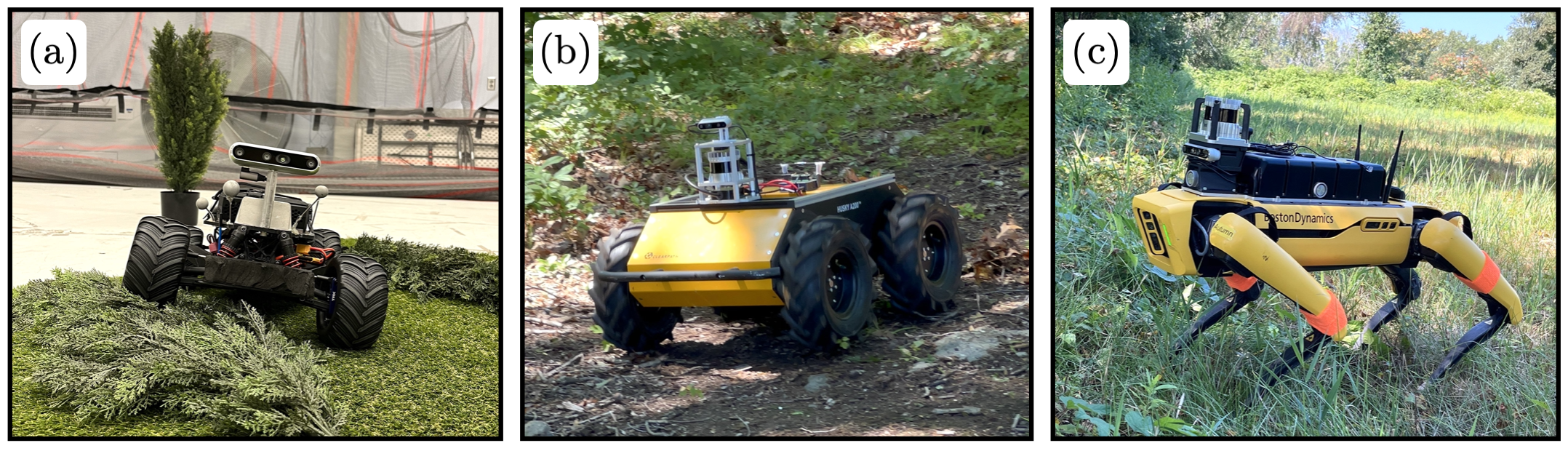}
\caption{Example ground robots that can be modeled with unicycle or bicycle dynamics models. (a) RC car. (b) Differential-drive robot. (c) Legged robot.}
\label{fig:hw_platforms}
\vspace*{-0.2in}
\end{figure}

Consider the discrete time system:
\begin{equation}\label{eq:dynamics}
    \mathbf{x}_{t+1} = F(\mathbf{x}_t, \mathbf{u}_t, \Param_{t})
\end{equation}
where $\mathbf{x}_t\in\setX\subseteq\R^n$ is the state vector such as the position and orientation of the robot, $\mathbf{u}_t\in\R^m$ is the control input provided to the robot, and $\Param_{t}\in\setParam\subseteq \R^r$ is the parameter vector that \edit{captures terrain traction.} 
We consider two models that are useful approximations of the dynamics of a wide range of robots as shown in Fig.~\ref{fig:hw_platforms}. 
Applicable to both differential-drive and legged robots, the \textbf{unicycle model} is defined as:
\begin{equation}\label{eq:unicycle}
    \begin{bmatrix}
        p_{t+1}^{x}\\
        p_{t+1}^{y}\\
        \theta_{t+1}
    \end{bmatrix} = 
    \begin{bmatrix}
        p_{t}^{x}\\
        p_{t}^{y}\\
        \theta_{t}
    \end{bmatrix} + \Delta \cdot
    \begin{bmatrix}
        \edit{\psi_{1, t}} \cdot v_t \cdot \sin{(\theta_t)}\\
        \edit{\psi_{1, t}} \cdot v_t \cdot \cos{(\theta_t)}\\
        \edit{\psi_{2, t}} \cdot \omega_t
    \end{bmatrix}
\end{equation}
where $\mathbf{x}_t=[p_{t}^{x}, p_{t}^{y}, \theta_{t}]\tr$ contains the X, Y positions and yaw, $\mathbf{u}_t=[v_t, \omega_t]\tr$ contains the commanded linear and angular velocities, \edit{$\Param_{t}=[\psi_{1,t}, \psi_{2, t}]\tr$} contains the linear and angular traction values \edit{$0\leq \psi_{1,t}, \psi_{2,t} \leq1$}, and $\Delta>0$ is the time interval. Intuitively, traction captures the ``slip'', or the \edit{ratio between achieved and commanded velocities.}
The \textbf{bicycle model} is applicable for Ackermann-steering robots and is defined as:
\begin{equation}\label{eq:bicycle}
    \begin{bmatrix}
        p_{t+1}^{x}\\
        p_{t+1}^{y}\\
        \theta_{t+1}%
    \end{bmatrix} = 
    \begin{bmatrix}
        p_{t}^{x}\\
        p_{t}^{y}\\
        \theta_{t}%
    \end{bmatrix} + \Delta \cdot
    \begin{bmatrix}
        \edit{\psi_{1, t}} \cdot v_t \cdot \cos{(\theta_t)}\\
        \edit{\psi_{1, t}} \cdot v_t \cdot \sin{(\theta_t)}\\
        \edit{\psi_{2, t}} \cdot v_t \cdot \tan (\delta_t)/L%
    \end{bmatrix}
\end{equation}
where $L$ is the wheelbase, $\mathbf{u}_t=[v_t, \delta_t]\tr$ contains the commanded linear velocity and steering angle, and $\bm{\psi}_t$ plays the same role as in the unicycle model. The reference point for the bicycle model in \eqref{eq:bicycle} is located at the center between the two rear wheels.

\subsection{\edit{Planning Objective}}
As this work focuses on fast navigation to the goal, we adopt the minimum-time objective used in~\cite{cai2022risk}, but other objectives \edit{for goal reaching} could also be used. Intuitively, the objective only assigns stage costs by accumulating the elapsed time before any state falls in the goal region. If the state trajectory does not intersect the goal region, the terminal cost further penalizes the estimated time-to-goal. 
\edit{Given a function $C^{\text{dist}}(\mathbf{x}_t)$ that measures the Euclidean distance between $\mathbf{x}_t$ and the goal, the minimum-time objective is defined over a state trajectory $\mathbf{x}_{0:T}$ from time $0$ to $T$:
\begin{align}
C(\mathbf{x}_{0:T}) \defeq C^{\text{term}}(\mathbf{x}_{0:T})+\sum_{t=0}^{T-1} C^{\text{stage}}(\mathbf{x}_{0:t})\label{eq:nominal_obj}
\end{align}
}
where the total cost consists of terminal and stage costs:
\begin{align}
\edit{C^{\text{term}}(\mathbf{x}_{0:T}) }&= \frac{C^{\text{dist}}(\mathbf{x}_T)}{s^{\text{default}}}\left( 1-\done{\mathbf{x}_{0:T}} \right) \\
\edit{C^{\text{stage}}(\mathbf{x}_{0:t}) }&= \Delta \left( 1-\done{\mathbf{x}_{0:t}} \right)\label{eq:stage_cost}
\end{align}
where $s^{\text{default}}>0$ is the default speed for estimating time-to-go and $\Delta>0$ is the constant time interval.
To avoid accumulating costs after arrival at the goal, we use an indicator function $\done{\mathbf{x}_{0:t}}$ that equals $1$ if any state in $\mathbf{x}_{0:t}$ has reached the goal, and equals $0$ otherwise. Although $\Delta$ is a constant, the number of time steps required to reach the goal changes according to the robot speed. Intuitively, this objective encourages the robot to reach the goal as quickly as possible.

\subsection{Key Challenges}
While the \edit{objective}~\eqref{eq:nominal_obj} can be optimized \edit{by finding an optimal control sequence} via nonlinear optimization techniques such as Model Predictive Path Integral control (MPPI~\cite[Algorithm~2]{williams2017information}), the terrain traction varies across terrain types and needs to be learned from real-world data. However, real-world terrain traction is uncertain since visually and geometrically similar terrain may have different traction properties (aleatoric uncertainty), and the traction models can only be trained on limited data (epistemic uncertainty). Even if uncertainty in terrain traction is quantified accurately, designing risk-aware planners that mitigate the risk of failures under this uncertainty is still challenging.
To address these challenges, we introduce our proposed uncertainty-aware traversability model and the risk-aware planner in Sec.~\ref{sec:traversability} and Sec.~\ref{sec:planning}, respectively. 

\section{Uncertainty-Aware Traversability Model}\label{sec:traversability}
\begin{figure*}[t]
\centering
\includegraphics[width=\linewidth, trim={0cm 0cm 0cm 0cm},clip]{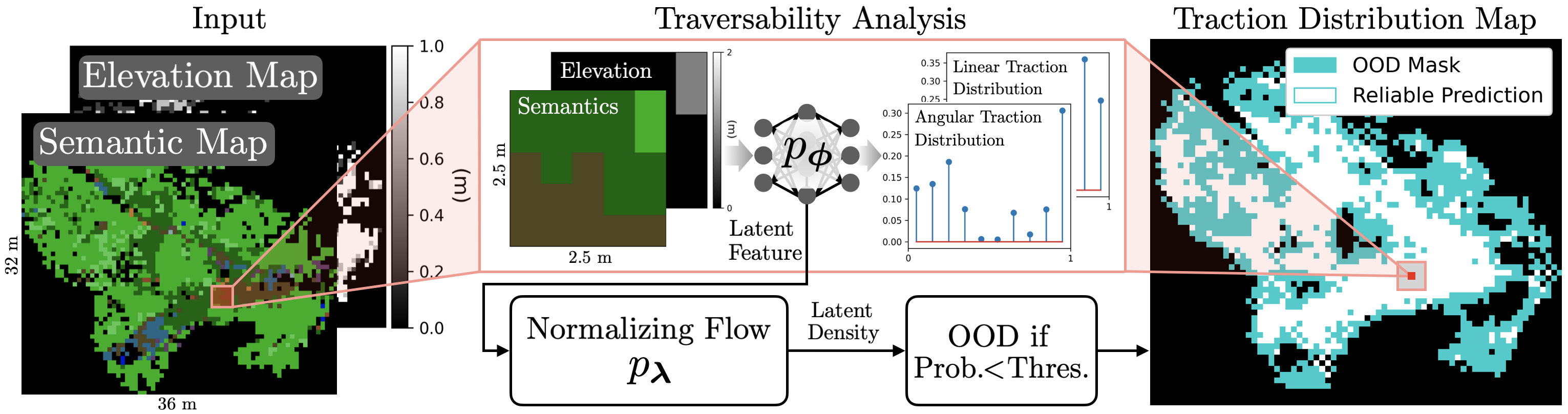}
\caption{
The proposed traversability pipeline maps elevation and semantic features to traction distributions that capture aleatoric uncertainty, and density for latent features that capture epistemic uncertainty. Terrain regions are deemed out-of-distribution (OOD) and later avoided during planning if the densities for the latent features are below a threshold. When the densities for latent features are above the threshold, the predicted traction distributions are reliable and inform downstream risk-aware planners (Sec.~\ref{sec:planning}) to trade off the risk of immobilization with the time savings by traversing regions with uncertain traction.}\label{fig:traversability_analysis_pipeline} 
\centering
\includegraphics[width=\linewidth, trim={0cm 0cm 0cm 0cm},clip]{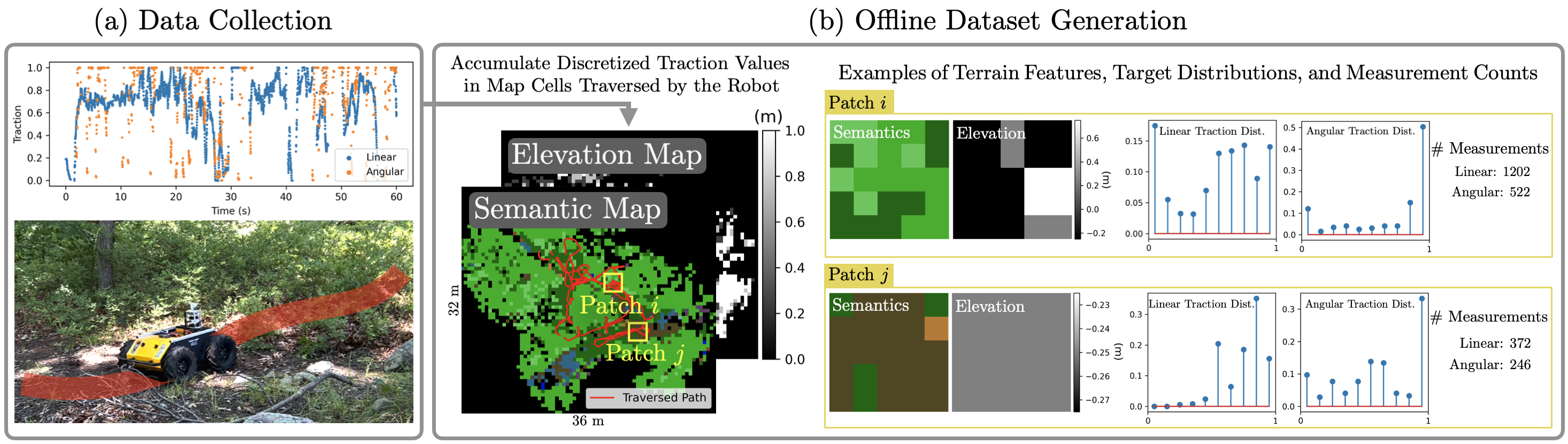}
\caption{
\edit{
Data collection and offline dataset generation.
(a)~An example of real world data collection using a Clearpath Husky. The robot was manually driven for 10 minutes while recording the traversed path, traction values and building semantic and elevation maps of the environment. The traction values were recorded at 20~Hz and only a subset of the collected traction values are shown for clarity, where the discontinuity in traction values occurred when linear or angular commands were not sent.
(b)~During offline dataset generation, traction values were discretized and accumulated via histograms stored in traversed map cells. The input to the traction predictor consists of semantic and elevation patches. Example terrain types include vegetation (light green), grass (dark green), dirt (light brown) and mulch (dark brown). The predicted and the empirical traction distributions are used to compute the training loss, and the associated measurement counts used to obtain the empirical traction distributions can be used to weight the training loss to discount rarely visited terrain.
}
}\label{fig:data_collection}
\vspace*{-0.2in}
\end{figure*}

In this section, we first introduce the traction distribution predictor that captures aleatoric uncertainty, and the latent space density estimator that captures epistemic uncertainty. An overview of the traversability analysis pipeline is shown in Fig.~\ref{fig:traversability_analysis_pipeline}. 
Then, we review \edit{the evidential method proposed by~\cite{natpn} in the context of traction learning, and propose a new uncertainty-aware loss to improve learning performance.}

\subsection{Aleatoric Uncertainty Captured in Traction Distribution}\label{sec:seq_train_aleatoric}

\edit{
Let $\setParam = \{\bm{\psi}^1, \dots, \bm{\psi}^B\}$ be a set of $B>0$ discretized traction values (ratios between achieved and commanded velocities), and $\setO$ be a set of terrain features containing elevation values and one-hot vectors 
of semantic labels. We want to model the distribution over $\setParam$ given an input terrain feature vector $\mathbf{o}\in\setO$:}
\begin{equation}
p_{\bm{\phi}}(\mathbf{o}):\setO \rightarrow \R_{\geq 0}^{B} \label{eq:cond_distribution}
\end{equation}
We use $\cat(p_{\bm{\phi}}(\mathbf{o}))$ to denote a categorical distribution over $\setParam$ 
\edit{that captures the aleatoric uncertainty due to environment factors that affect traction but are not captured in the terrain features~$\mathbf{o}$.} 
Note that ~\eqref{eq:cond_distribution} can be learned by an NN parameterized by $\bm{\phi}$ that can be trained using an empirically collected dataset $\{(\mathbf{o}, \Param)_k\}_{k=1}^{K}$ where $K>0$.
\edit{While we do not explicitly account for the uncertainty in the terrain features (e.g., noisy elevation estimates, or misclassification of terrain types due to visual similarity) or other factors such as the design of the low-level velocity controller, these unmodeled effects will manifest in the empirical dataset and can still be implicitly captured in the learned traction distributions.}

\edit{
We use categorical distributions as convenient alternatives to Gaussian Mixture Models and normalizing flows~\cite{Kobyzev2021flow} for learning multi-modal traction distributions observed in practice, because they do not require tuning the number of clusters, generate bounded distributions by construction, and converge faster than normalizing flows based on our empirical experience while achieving similar accuracy.
Generally, discretizing a high-dimensional space can be challenging as the number of bins grows exponentially with the dimension. However, categorical distributions are well-suited in our case since we only need to discretize 1-D linear and angular traction values. Therefore a relatively small number of discrete bins suffice.
}

Examples of real world data collection and offline dataset generation can be found in Fig.~\ref{fig:data_collection}. The semantic and geometric information about the environment can be built by using a semantic octomap~\cite{asgharivaskasi2021active} that temporally fuses semantic point clouds. 
\edit{
We used PointRend~\cite{kirillov2020pointrend} trained on the RUGD off-road navigation dataset~\cite{RUGD2019IROS}
with 24 semantic categories to segment RGB images and subsequently projected the semantics onto lidar point clouds. 
During offline dataset generation, we obtained the empirical linear and angular traction distributions by accumulating discretized traction measurements in histograms stored in every terrain cell traversed by the robot. The measurement counts were also stored so that, during training, we could weight the loss for each cell by the measurement counts to discount rarely visited terrain.
In practice, we learned the linear and angular traction distributions separately. We used a shared encoder (convolutional layers followed by fully connected layers) to process the semantic and elevation patches of the terrain. The shared encoder is followed by two fully connected decoder heads with soft-max outputs for predicting the linear and angular traction distributions.}

\subsection{Epistemic Uncertainty Captured in Latent Space Density}\label{sec:seq_train_epistemic}
Due to limited training data, the predicted traction distributions for novel parts of the terrain may be unreliable and lead to degraded navigation performance in those regions. 
To measure epistemic uncertainty, we want to estimate the density of the latent feature 
\edit{$\mathbf{z}^{\mathbf{o}}\in\R^H$}
obtained from an intermediate layer of the traction predictor $p_{\bm{\phi}}$~\eqref{eq:cond_distribution} based on the terrain feature $\mathbf{o}$. The density estimator is defined as:
\begin{equation}
    p_{\bm{\lambda}}(  \edit{ \mathbf{z}^{\mathbf{o}} }   ): \R^H \rightarrow \R_{\geq 0}\label{eq:p_lambda_latent_density}
\end{equation}
where we use a normalizing flow parameterized by $\bm{\lambda}$ to learn~\eqref{eq:p_lambda_latent_density}.
At a high level, a normalizing flow works by transforming an arbitrary target distribution into a simple base distribution such as a standard normal via a sequence of invertible and differentiable mappings. Then, the density of a sample $\mathbf{z}^{\mathbf{o}}$ can be computed by change of variable formula~\cite{Kobyzev2021flow} --- it is the product of the density of the transformed sample under the base distribution, and the change in \edit{volume} measured by the determinant of the Jacobian of the transformation.
\edit{
When selecting the latent space features, it is crucial to ensure that they contain task-relevant information. To this end, we use the latent features produced by the shared terrain feature encoder, because they contain information useful for predicting both linear and angular traction distributions.
}

\edit{
For interpretation purposes,
}
we design a simple confidence score \edit{$g(\mathbf{z}^\mathbf{o})$} for input feature $\mathbf{o}$ based on the maximum density $p^{\text{max}}\in\R_{\geq0}$ and minimum density $p^{\text{min}}\in\R_{\geq0}$ observed for the latent features of terrain in the training dataset: 
\begin{equation}\label{eq:latent_conf}
\edit{ g(\mathbf{z}^\mathbf{o}) } = \frac{p_{\bm{\lambda}}(  \edit{ \mathbf{z}^{\mathbf{o}} }  ) - p^{\text{min}}}{p^{\text{max}}-p^{\text{min}}}.
\end{equation}
\edit{
During deployment, terrain features with a confidence score below some threshold $g^{\text{thres}}\in\R$ are deemed OOD; these regions with  OOD terrain features can be explicitly avoided during planning via auxiliary penalties. A principled way to set $g^{\text{thres}}$ is to use the $\kappa$-th percentile of the densities obtained from all the terrain features in the training dataset, where a higher value of $\kappa\in[0,100]$ will cause more terrain features to be classified as OOD at test time. Because of the normalization in~\eqref{eq:latent_conf}, $g^{\text{thres}}=0$ and $g^{\text{thres}}=1$ conveniently correspond to the {0-th} and {100-th} percentiles. Note that the threshold can be selected offline, and $g^{\text{thres}}=0$ can be used if the robot should only avoid terrain features with densities lower than densities observed during training.
This strategy improves navigation success rate when the learned traction models are deployed in environments unseen during training, both in simulations (see Sec.~\ref{sec:exp_bag_sim}) and in hardware experiments (see Sec.~\ref{sec:results:outdoor_exp_spot}).
}

\subsection{Evidential Deep Learning}\label{sec:traversability:joint_training_natpn_uce}
\edit{
While the traction predictor and the density estimator can be trained sequentially, Charpentier et. al.~\cite{natpn} have shown that joint training using evidential deep learning can improve OOD detection performance while retaining prediction accuracy. In this section, we review the method and training loss proposed in~\cite{natpn}, where NN outputs parameterize Dirichlet distributions (the conjugate priors of categorical distributions).
}

\edit{
The Dirichlet distribution $q=\dir(\bm{\beta})$ with concentration parameters $\bm{\beta}=[\beta_1,\dots,\beta_B]\tr \in\R^{B}_{> 0}$  is a hierarchical distribution over categorical distributions $\cat(\mathbf{p})$, where $\mathbf{p}\in\R^{B}_{\geq0}$ is a normalized probability mass function (PMF) over $B>0$ bins, i.e. $\sum_{b=1}^B p_b = 1$. The parameters $\mathbf{p}$ of the lower level categorical distribution $\mathrm{Cat}(\mathbf{p})$ are sampled from the higher level Dirichlet distribution i.e. $\mathbf{p}\sim\dir(\bm{\beta})$. The mean (also called the \textit{expected PMF}) of the Dirichlet distribution is given by $\mathbb{E}_{\mathbf{p} \sim q} \left[ \mathbf{p} \right] = \bm{\beta}/\sum_{b=1}^{B} \beta_b$. The expected PMF captures aleatoric uncertainty.
The sum of the parameters $\bm\beta$ i.e. $\sum_{b=1}^{B} \beta_b$ represents how concentrated the Dirichlet distribution is around its mean. Therefore, $\sum_{b=1}^{B} \beta_b$ is also known as the concentration parameter, and corresponds to the ``total evidence" of a data point observed in the training set. Higher data evidence corresponds to lower epistemic uncertainty.
Given a prior Dirichlet belief $\dir(\bm{\beta}^{\text{prior}})$ and the input feature $\mathbf{o}$, the NN performs an input-dependent posterior update: 
\begin{align}
\bm{\beta}_{\bm{\phi}, \bm{\lambda}}^{\mathbf{o}} &= \bm{\beta}^{\text{prior}} + n_{\bm{\lambda}}^{\mathbf{o}} p_{\bm{\phi}}(\mathbf{o})\label{eq:dirichlet_posterior} \\
n_{\bm{\lambda}}^{\mathbf{o}} &=Np_{\bm{\lambda}}(\mathbf{z}_{\mathbf{o}})\label{eq:dir_new_evidence}
\end{align}
where the posterior Dirichlet distribution $q_{\bm{\phi}, \bm{\lambda}}^{\mathbf{o}} = \dir(\bm{\beta}_{\bm{\phi}, \bm{\lambda}}^{\mathbf{o}})$ depends on the predicted traction $p_{\bm{\phi}}(\mathbf{o})$~\eqref{eq:cond_distribution} and the predicted evidence $n_{\bm{\lambda}}^{\mathbf{o}}$ that is proportional to the density for the latent feature $p_{\bm{\lambda}}(\mathbf{z}_{\mathbf{o}})$~\eqref{eq:p_lambda_latent_density} weighted by a fixed certainty budget $N>0$.
The posterior Dirichlet distribution leads to the expected traction PMF:
\begin{equation}\label{eq:dirichlet_posterior_mean}
    \mathbf{p}^{\mathbf{o}}_{\bm{\phi}, \bm{\lambda}} = \frac{n^\text{prior}\mathbf{p}^\text{prior}+ n_{\bm{\lambda}}^{\mathbf{o}} p_{\bm{\phi}}(\mathbf{o}) }{n^\text{prior}+  n_{\bm{\lambda}}^{\mathbf{o}} }
\end{equation}
where  $n^\text{prior}=\sum_{b=1}^{B} \beta_b^\text{prior}$ and $\mathbf{p}^{\text{prior}}=\bm{\beta}^{\text{prior}}/n^{\text{prior}}$. We use a flat prior by setting $\bm{\beta}^{\text{prior}}=\mathbf{1}_{B}$, where $\mathbf{1}_{B}\in\R^{B}$ is a vector of all ones, such that $\dir(\bm{\beta}^\mathrm{prior})$ is a uniform distribution over all PMFs.
Based on this formulation proposed by~\cite{natpn}, the posterior Dirichlet distribution $q_{\bm{\phi}, \bm{\lambda}}^{\mathbf{o}}$ and expected traction distribution $\mathbf{p}^{\mathbf{o}}_{\bm{\phi}, \bm{\lambda}}$ both depend on the traction predictor, density estimator and the input features. 
While the analysis of loss functions we perform below is for a generic Dirichlet distribution $q=\dir(\bm{\beta})$ and PMF $\mathbf{p}$ for notational convenience, the posterior Dirichlet distribution and its expected PMF should be substituted by (\ref{eq:dirichlet_posterior},~\ref{eq:dir_new_evidence},~\ref{eq:dirichlet_posterior_mean}) during training. 
}

\edit{
Given the target PMF $\mathbf{y}\in\R^B_{\geq0}$ that contains the empirically estimated traction distribution, the traction predictor and the normalizing flow  can be trained jointly with the following uncertainty-aware cross entropy (UCE) loss~\cite{natpn}:
\begin{equation}\label{eq:uce_h}
    L^\text{UCE}(q, \mathbf{y}) - H(q)
\end{equation}
where $L^\text{UCE}(q, \mathbf{y}) \defeq \E_{\mathbf{p}\sim q} [- \sum_{b=1}^B y_b\log p_b ]$ is defined as the expected cross entropy loss and $H(q)$ is an entropy term that encourages smoothness of $q$. Note that both $L^\text{UCE}$ and $H(q)$ depend on $\bm{\beta}$ (see Appendix~\ref{appendix:uce_details} for details).
}

The ablation study in~\cite{natpn} has shown that \edit{training with~\eqref{eq:uce_h} improves OOD detection performance while retaining similar accuracy achieved using the conventional cross entropy (CE) loss.}
However, the key limitation of CE-based losses in our use case is that they treat the prediction errors across bins independently. The independence assumption is undesirable for learning traction where bins are obtained by discretizing continuous traction values. These bins are ordered --- bins closer to each other should be treated more similarly than bins far apart. We address this limitation by proposing a new loss function based on the squared Earth Mover's Distance~\cite{hou2016emd2} that has been shown to achieve better accuracy than $\ce$-based losses when bins are ordered.

\subsection{Uncertainty-Aware Squared Earth Mover's Distance}\label{sec:traversability:joint_training_natpn_uemd2}

\begin{figure}[t!]
	\centering
	\includegraphics[width=0.95\linewidth, trim={0.cm 0.cm 0.cm 0.cm},clip]{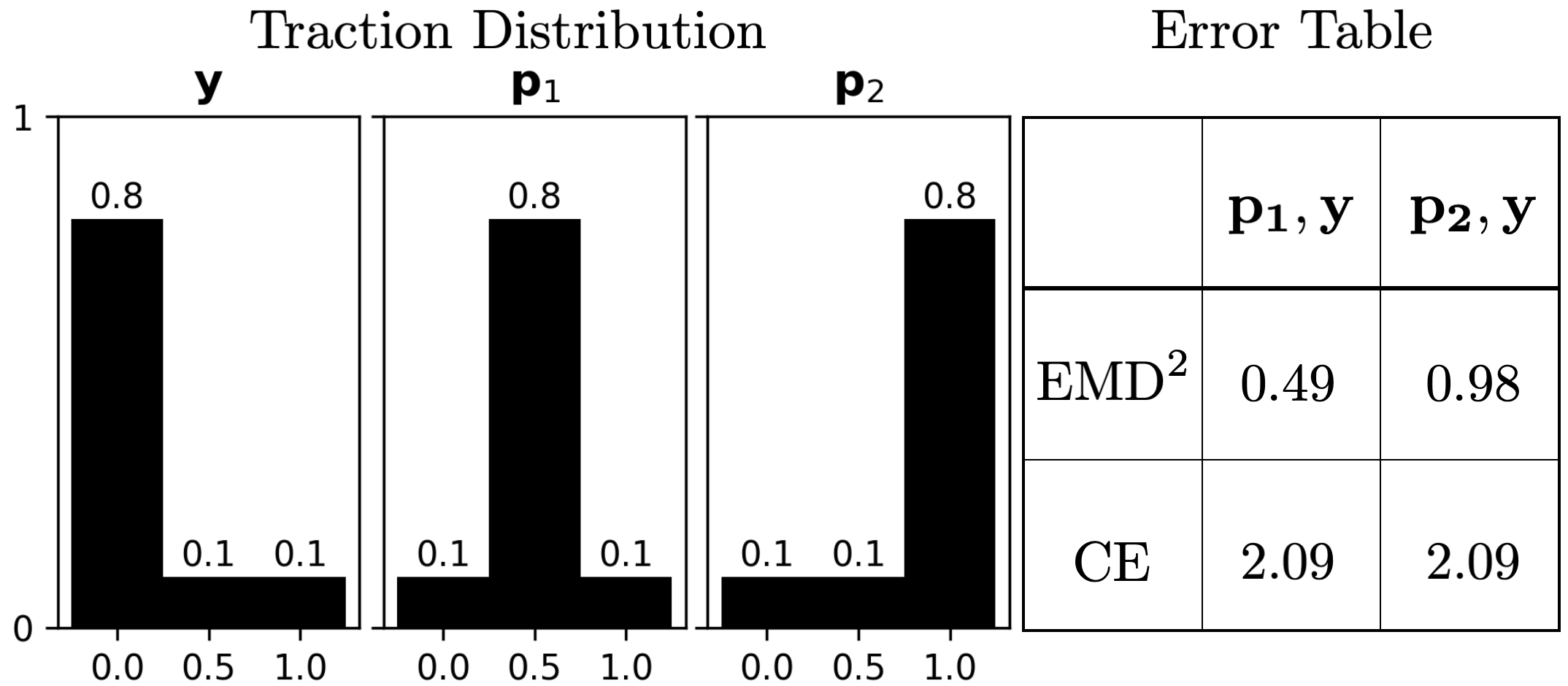}
	\caption{Difference between $\emdsq$ and CE. Given the ground truth $\mathbf{y}$ and the predictions $\mathbf{p}_1$ and $\mathbf{p}_2$, CE produces the same values while $\emdsq$ penalizes $\mathbf{p}_2$ more. In practice, $\emdsq$ is more desirable because it accounts for the cross-bin relationship among the discretized traction values.}
	\label{fig:example_emd2_vs_ce}
\vspace*{-0.2in}
\end{figure}

Intuitively the Earth Mover's Distance (EMD) between two distributions measures the minimum cost of transporting the probability mass of one distribution to the other, which has a closed-form solution for two categorical distributions defined by PMFs with the same number of bins~\cite{hou2016emd2}. Given a predicted PMF $\mathbf{p}\in\R^B_{\geq0}$ and the target $\mathbf{y}\in\R^B_{\geq0}$, the normalized EMD with $l$-norm for $B$ equally-spaced bins can be computed in closed-form~\cite{hou2016emd2}:
\begin{equation}
    \emd(\mathbf{p}, \mathbf{y}) = \Big(\frac{1}{B}\Big)^{\frac{1}{l}} \| \cumsum(\mathbf{p}) - \cumsum(\mathbf{y})  \|_l 
\end{equation}
where $\cumsum:\R^B\rightarrow \R^B$ is the cumulative sum operator.  For convenience during training, we use $l=2$ for Euclidean distance and optimize the squared EMD loss ($\emdsq$), dropping the constant factor. The toy example in Fig.~\ref{fig:example_emd2_vs_ce} clearly shows that $\emdsq$ better captures the physical meaning of the predicted PMFs than CE which ignores the relationship between bins.

\begin{figure*}[t!]
	\newcommand{\y}{{\color{red}\mathbf{y}}}
	\centering
	\includegraphics[width=\linewidth, trim={0.cm 0.cm 0.cm 0.cm},clip]{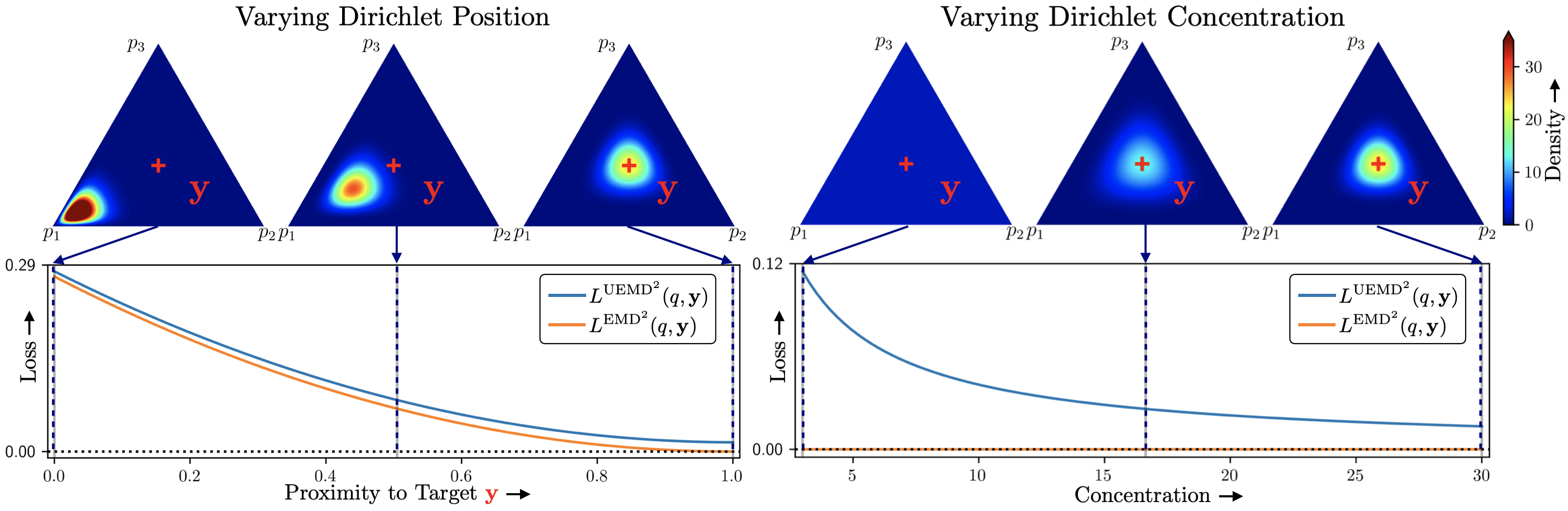}
	\caption{
		Analyzing the difference between the standard $\emdsq$ loss and our proposed $\uemdsq$ loss on a toy example with three bins $p_1, p_2, p_3$.
		Each blue triangle represents a predicted Dirichlet distribution $q$ visualized as a probability density over the 3-simplex; each point inside the simplex corresponds a categorical distribution over the three bins. 
\edit{The red cross ${\color{red}\textbf{+}}$ denotes the location of the target label distribution $\y$ in the training set.}
  A Dirichlet distribution can be parametrized by two quantities: the position of its mean and the concentration around its mean.
		\textbf{\textit{Left:}} varying the position of the Dirichlet while keeping its concentration fixed. In this case, both losses behave similarly and as desired---they encourage the predicted Dirichlet to be close to the target label distribution.
		\textbf{\textit{Right:}} varying the concentration of the Dirichlet while keeping its position fixed to the ground truth.
		Since $\emdsq$ only depends on the position of the Dirichlet mean, it is constant with respect to varying concentration.
		However, our proposed $\uemdsq$ encourages the predicted Dirichlet to have high concentration (low epistemic uncertainty).
		Learning to predict low epistemic uncertainty for in-distribution training examples is essential for calibrated uncertainty prediction and detecting out-of-distribution examples, as opposed to being indifferent to the concentration. 
	}
\label{fig:example_emd2_vs_uemd2}
\vspace*{-0.2in}
\end{figure*}

As $\emdsq$ is only defined for PMFs, a na\"ive strategy is to compare the target $\mathbf{y}$ to the expected PMF from the predicted Dirichlet $q$, which leads to the following loss function (ignoring the constant multiplicative term):
\begin{align}
 L^{\emdsq}(q, \mathbf{y}) &\defeq \| \cumsum(\overline{\mathbf{p}}) - \cumsum(\mathbf{y})  \|_2^2 \nonumber \\
 &\ = \cumsum(\overline{\mathbf{p}})\tr \cumsum(\overline{\mathbf{p}}) + \eta(q, \mathbf{y}) \label{eq:emd2}
\end{align}
where $\overline{\mathbf{p}}\defeq E_{\mathbf{p}\sim q }[\mathbf{p}] = \bm{\beta}/\beta_0$ is the expected PMF, \edit{$\beta_0\defeq \sum_{b=1}^B \beta_b$ is the total evidence} and
\begin{equation}\label{eq:emd_l_term}
     \eta(q, \mathbf{y}) \defeq - 2\ \cumsum(\overline{\mathbf{p}})\tr\cumsum(\mathbf{y}) + \cumsum(\mathbf{y})\tr\cumsum(\mathbf{y}).
\end{equation}
Note that $\cumsum(\overline{\mathbf{p}})$ can also be written as $\cumsum(\bm{\beta}) / \beta_0$ due to the linearity of the cumulative sum operator. 
However, $L^{\emdsq}$ is invariant to the \edit{total evidence $\beta_0$} of the Dirichlet distribution, as illustrated in the toy example in Fig.~\ref{fig:example_emd2_vs_uemd2}, so the epistemic uncertainty cannot be learned accurately.

Similar to the approach in~\cite{natpn} that uses the expectation of the cross entropy loss that depends on $\bm{\beta}$, we propose the uncertainty-aware $\emdsq$ ($\uemdsq$) loss defined as the expectation of the $\emdsq$ given the Dirichlet $q$:
\begin{align}
L^{\uemdsq}(q, \mathbf{y}) \defeq \E_{\mathbf{p}\sim q}~\big[ \emdsq (\mathbf{p}, \mathbf{y}) \big].\label{eq:uemd2_def}
\end{align}
The following theorem states that our proposed $\uemdsq$ loss can be computed in a closed form.
\begin{theorem}\label{prop:uemd2}
\edit{
Let $q = \dir(\bm{\beta})$ be a Dirichlet distribution and let $\cat(\mathbf{y})$ be a categorical distribution.
Then, a closed-form expression exists for $L^{\uemdsq}(q, \mathbf{y})$ given by:
\begin{align}
L^{\uemdsq} \big(q, \mathbf{y}\big) = \cumsum(\overline{\mathbf{p}})\tr \frac{\cumsum(\bm{\beta})+\mathbf{1}_B}{\beta_0+1} + \eta(q, \mathbf{y})\label{eq:uemd2}
\end{align}
where $\overline{\mathbf{p}}=\E_{\mathbf{p}\sim q}[\mathbf{p}]$, and $\eta$ is defined in~\eqref{eq:emd_l_term}.
}
\end{theorem}
\begin{proof}
See Appendix~\ref{appendix:uemd2_proof}.
\end{proof}

Due to structural similarity with $L^{\emdsq}$~\eqref{eq:emd2}, the proposed loss~\eqref{eq:uemd2} also penalizes the $\emdsq$ error to encourage accurate traction predictions. In addition, the proposed loss penalizes low concentration $\bm{\beta}$ to encourage low epistemic uncertainty as shown in Fig.~\ref{fig:example_emd2_vs_uemd2}. In fact, it can be proved that $L^{\uemdsq}$ is always greater or equal to $L^{\emdsq}$ using Jensen's inequality and the convexity\footnote{Taking cumulative sum and squared difference are convex operations.} of $L^{\emdsq}$. 
\edit{
While~\eqref{eq:uemd2} can be directly used as a loss function, EMD$^2$-based loss may not always converge to the desired local optima as observed by~\cite{hou2016emd2}. To address this issue, we follow~\cite{hou2016emd2} by considering a loss function that combines both EMD$^2$-based and CE-based loss terms. Therefore, we consider the following multi-objective optimization:
\begin{equation}\label{eq:uce_uemd2_h}
    L^{\uce}(q, \mathbf{y}) + w_1 L^{\uemdsq}(q, \mathbf{y}) - w_2 H(q)
\end{equation}
where the entropy $H(q)$ encourages smoothness and the weights $w_1, w_2 \geq 0$ are hyperparameters.
In practice, we compute~\eqref{eq:uce_uemd2_h} for the predicted linear and angular traction distributions separately and average the loss values.}
As simulation results suggest in Sec.\ref{sec:prediction_accuracy_and_ood_detection}, the multi-objective loss~\eqref{eq:uce_uemd2_h} leads to more stable training and \edit{better generalization to test data.}

\section{Planning with Learned Traction Distribution}\label{sec:planning}
\edit{
While OOD terrain causing high epistemic uncertainty should be avoided, in-distribution terrain may still lead to high aleatoric uncertainty due to complex vehicle-terrain interactions. Therefore, we propose a risk-aware planner that trades off the risk of immobilization with potential time savings from traversing terrain that leads to high aleatoric uncertainty.
}

\subsection{Conditional Value at Risk (CVaR)}

\begin{figure}[t]
	\centering
	\includegraphics[width=0.7\linewidth, trim={0cm 0cm 0cm 0cm},clip]{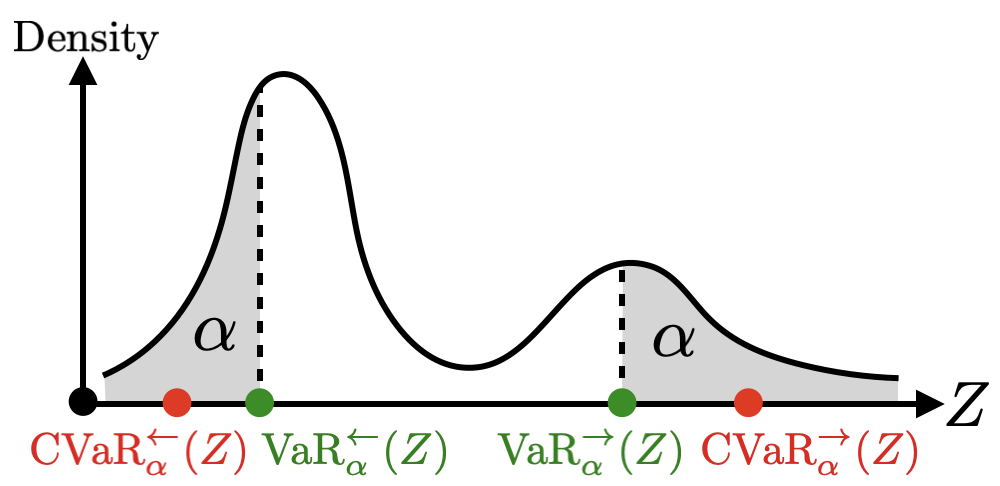}
	\caption{This work defines two versions of Conditional Value at Risk (CVaR) to capture the worst-case expected values at either the left tail as $\lcvar{\alpha}{Z}$ or the right tail as $\rcvar{\alpha}{Z}$ for some random variable $Z$, where the worst-case scenarios constitute $\alpha\in(0,1]$ portion of total probability. The left-tail and right-tail Values at Risk (VaR) are defined as $\lvar{\alpha}{Z}$ and $\rvar{\alpha}{Z}$.}
	\label{fig:cvar_definition}
\vspace*{-0.2in}
\end{figure}

We adopt the Conditional Value at Risk (CVaR) as a risk metric because it satisfies a group of axioms important for rational risk assessment~\cite{majumdar2020should}. The conventional definition of CVaR assumes the worst-case occurs at the right tail of the distribution. We define CVaR for a random variable $Z$ at level $\alpha\in(0,1]$ for both the right and left tails of its distribution (see Fig.~\ref{fig:cvar_definition}) as follows:
\begin{align}
    \rcvar{\alpha}{Z} &\defeq \frac{1}{\alpha} \int_{0}^{\alpha} \rvar{\tau}{Z}\ d\tau \\
    \lcvar{\alpha}{Z} &\defeq \frac{1}{\alpha} \int_{0}^{\alpha} \lvar{\tau}{Z}\ d\tau
\end{align}
where the right and left Values at Risk (VaR) are defined as:
\begin{align}
\rvar{\alpha}{Z} &\defeq \min  \{ z \mid p(Z > z) \leq \alpha\} \\
\lvar{\alpha}{Z} &\defeq \max \{ z \mid p(Z < z) \leq \alpha\}.
\end{align}
Intuitively, $\rcvar{\alpha}{Z}$ and $\lcvar{\alpha}{Z}$ capture the expected outcomes that fall in the right tail and left tail of the distribution, respectively, where each tail occupies $\alpha$ portion of the total probability. Note that the right-tail definitions are suitable for costs to be minimized, and the left-tail definitions are suitable for low traction values. When $\alpha=1$, either definition of CVaR is equivalent to the mean of the distribution $\E [Z]$.

\subsection{\edit{Risk-Aware Planning}}\label{sec:risk_aware_cost}
To account for the risk due to uncertain traction, we first present an existing approach~\cite{wang2021adaptive} that optimizes for the right-tail CVaR of the planning objective (\cvarcost{}), and then propose a more computationally efficient method that accounts for the left-tail CVaR of traction (\cvardyn{}). \edit{Lastly, we discuss the advantages and limitations of these two methods.}

\subsubsection{Worst-Case Expected Cost (\cvarcost{}~\cite{wang2021adaptive})}

\edit{Given the initial state $\mathbf{x}_0$, we want to find a control sequence $\mathbf{u}_{0:T-1}$ that minimizes the worst-case expected value of the nominal objective $C$~\eqref{eq:nominal_obj} given uncertain terrain traction:}
\edit{
\begin{align}
\min_{\mathbf{u}_{0:T-1}} \quad & \rcvar{\alpha}{C(\widetilde{\mathbf{x}}_{0:T})} \label{eq:cvar_cost}\\
\textrm{s.t.} \quad & \widetilde{\mathbf{x}}_{t+1} = F(\widetilde{\mathbf{x}}_t, \mathbf{u}_t, \widetilde{\Param}_{t})\\
&\widetilde{\Param}_t \sim \cat (\mathbf{p}^{\widetilde{\mathbf{o}}_t}_{\bm{\phi}, \bm{\lambda}})\\
&\widetilde{\mathbf{o}}_t\text{ is the terrain feature at }\widetilde{\mathbf{x}}_t\\
&\widetilde{\mathbf{x}}_0=\mathbf{x}_0 \quad \forall t\in\{0,\dots,T-1\}
\end{align}
where traction $\widetilde{\Param}_t$ is realized based on the predicted traction PMF $\mathbf{p}^{\widetilde{\mathbf{o}}_t}_{\bm{\phi}, \bm{\lambda}}$~\eqref{eq:dirichlet_posterior_mean} after observing the terrain feature $\widetilde{\mathbf{o}}_t$. Due to the uncertain traction, the original objective $C(\widetilde{\mathbf{x}}_{0:T})$ is now a random variable that depends on the realization of the state trajectory. Note that this approach is inspired by~\cite{wang2021adaptive}, but we additionally handle terrain-dependent traction distributions. 
}

\edit{
In practice, optimizing~\eqref{eq:cvar_cost} using MPPI requires a subroutine that empirically estimates the right-tail CVaR of the objective by collecting $M>0$ realizations of the nominal objective $\{C(\widetilde{\mathbf{x}}^m_{0:T})\}_{m=1}^M$ for each candidate control sequence by using sampled traction values. 
To exploit GPU parallelization, we pre-generate $M>0$ traction maps where each map cell contains sampled traction values.
As a result, each candidate control sequence can be evaluated in parallel for all the pre-generated traction maps. While the sampled traction maps can be reused, the computation can still grow prohibitively as the map size grows. Therefore, we propose a cheaper cost design that accounts for the left-tail CVaR of terrain traction.
}

\subsubsection{Worst-Case Expected Terrain Traction (\cvardyn{})}

\edit{
Given the initial state $\mathbf{x}_0$, we want to find a control sequence $\mathbf{u}_{0:T-1}$ that minimizes the nominal objective $C$~\eqref{eq:nominal_obj} using the state trajectory obtained with the worst-case expected traction:
\begin{align}
\min_{\mathbf{u}_{0:T-1}} \quad & C(\bar{\mathbf{x}}_{0:T}) \label{eq:cvar_dyn}\\
\textrm{s.t.} \quad & \bar{\mathbf{x}}_{t+1} = F(\bar{\mathbf{x}}_t, \mathbf{u}_t, \bar{\Param}_{t})\\
&\bar{\Param}_{t}=
    \begin{bmatrix}
    \lcvar{\alpha}{\bar{\psi}_{1,t}} \\
    \lcvar{\alpha}{\bar{\psi}_{2,t}}
    \end{bmatrix},\ 
    \begin{bmatrix}
    \bar{\psi}_{1,t} \\
    \bar{\psi}_{2,t}
    \end{bmatrix}\sim \cat ( \mathbf{p}^{ \bar{\mathbf{o}}_t }_{\bm{\phi}, \bm{\lambda}} ) \\
&\bar{\mathbf{o}}_t\text{ is the terrain feature at }\bar{\mathbf{x}}_t\\
&\bar{\mathbf{x}}_0=\mathbf{x}_0 \quad \forall t\in\{0,\dots,T-1\}
\end{align}
where $\bar{\Param}_{t}$ contains the left-tail CVaR of the linear and angular traction based on the predicted traction PMF $\mathbf{p}^{ \bar{\mathbf{o}}_t }_{\bm{\phi}, \bm{\lambda}}$~\eqref{eq:dirichlet_posterior_mean} after observing the terrain feature $\bar{\mathbf{o}}_t$. 
}
When $\alpha=1$, the expected values of the traction parameters are used, equivalent to the
\edit{planning approach used by~\cite{Gasparino2022wayfast}}.

\subsubsection{\edit{Advantages and Limitations}}\label{sec:subsec}

\edit{
Both \cvarcost{} and \cvardyn{} leverage intuitive notions of risk based on the worst-case expected cost and traction, respectively. Moreover, they are simple to tune with a single risk parameter $\alpha$ regardless of the number of terrain types.
Note that \cvarcost{} is a general algorithm that handles uncertainty in the planning problem~\cite{wang2021adaptive}.
In comparison, \cvardyn{} is computationally cheaper, but it exploits the intuition that low traction usually worsens time-to-goal. However, such relationship between system parameters and task performance may not hold for more complicated systems and different tasks.}

\section{Evaluation of Traversability Learning Pipeline}\label{sec:benchmark_traversability_learning_pipeline}
The proposed evidential traversability learning method is benchmarked using a synthetic terrain dataset (Sec.~\ref{sec:synthetic_3d_terrain}) designed to simulate data scarcity during real-world data collection and provide ground truth traction distributions and OOD terrain masks. Several variants of the proposed loss~\eqref{eq:uce_uemd2_h} are compared based on prediction accuracy and OOD detection performance (Sec.~\ref{sec:prediction_accuracy_and_ood_detection}). To highlight the benefits of joint training and our $\uemdsq$ loss~\eqref{eq:uemd2}, we provide an ablation study in Sec.~\ref{sec:uemd2_ablation}. After analyzing the proposed planner in Sec.~\ref{sec:benchmark_planners}, Sec.~\ref{sec:emd_and_nav_performance} contains key results that show improved navigation performance due to our proposed loss.

As comparing uncertainty quantification methods is not the main focus of this work, we refer interested readers to~\cite{natpn} that has demonstrated the computational advantages, learning accuracy and OOD detection performance of the NN architecture used in this work compared to the other state-of-the-art uncertainty quantification methods.

\subsection{Synthetic 3D Terrain Datasets}\label{sec:synthetic_3d_terrain}

\begin{figure*}[t]
\centering
\includegraphics[width=0.98\linewidth, trim={0.cm 0.cm 0.cm 0.cm},clip]{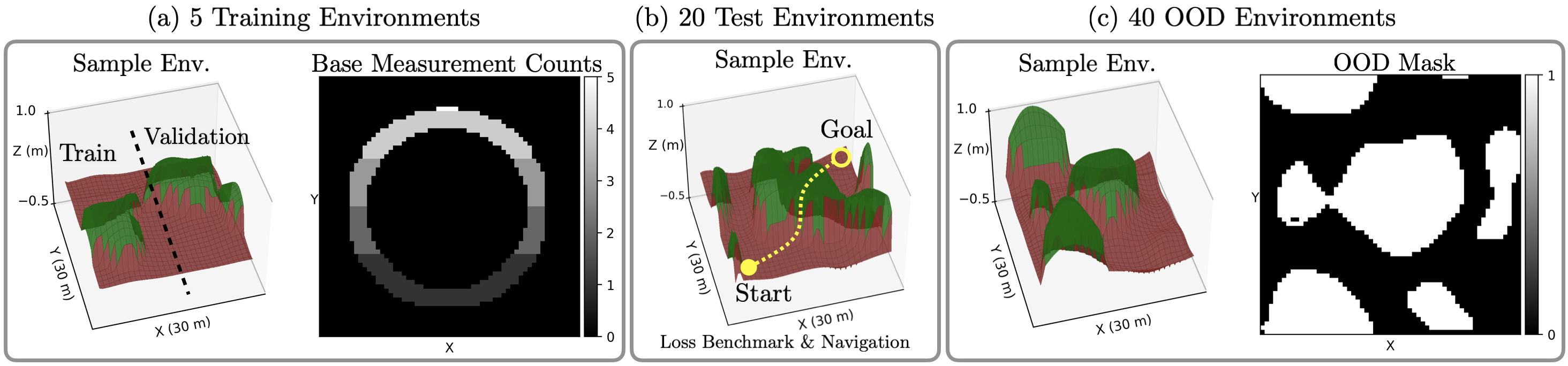}
\caption{
The synthetic 3D terrain dataset with dirt (brown) and vegetation (green) semantic types.
(a) In each training environment, there are limited traction measurements along a pre-specified circular path to mimic real-world data collection \edit{with limited coverage.} Each environment is split into two for cross validation. Furthermore, we analyze the effect of a varying number of measurements by multiplying the base measurement counts (see Fig.~\ref{fig:loss_function_benchmark}).
(b) The test environments contain novel terrain features for analyzing the traction prediction accuracy. To support the key argument that $\emdsq$ is a better indicator for navigation performance, models trained with different loss functions are deployed in the test environments for go-to-goal tasks (see Sec.~\ref{sec:emd_and_nav_performance}).
(c) Compared to the test environments, the OOD dataset additionally provides binary masks for the novel terrain with elevation and/or slope not observed during training.
}
\label{fig:synthetic_datasets}
\vspace*{0.1in}
\centering
\includegraphics[width=\linewidth, trim={0.cm 0.2cm 0.cm 0.2cm},clip]{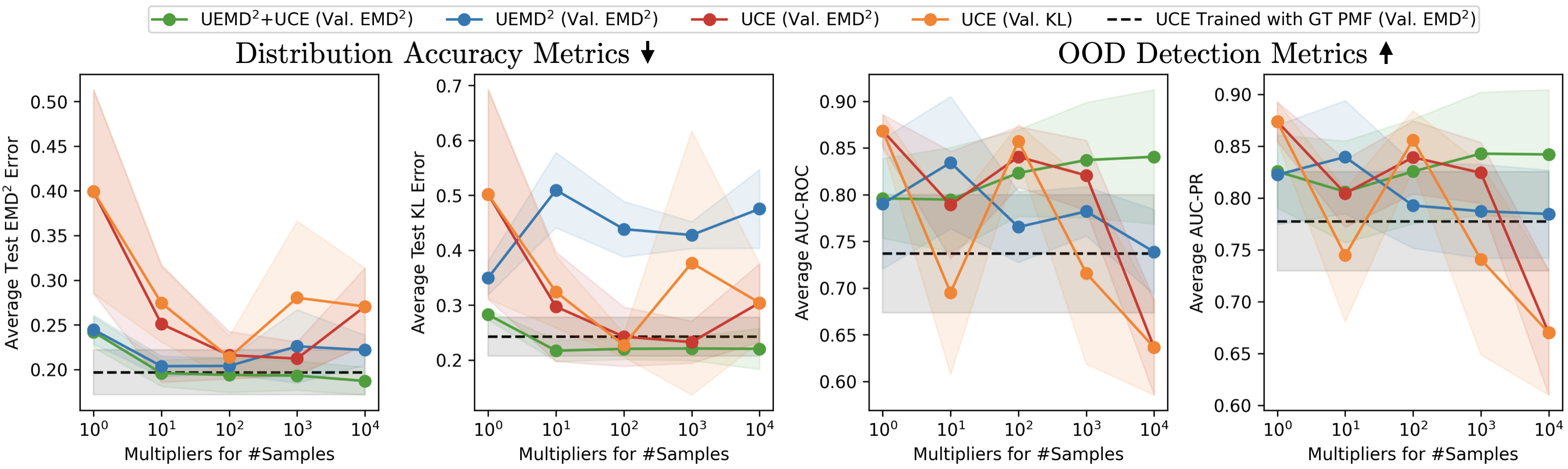}
\caption{Prediction errors measured in $\emdsq$ and KL divergence (the lower, the better), and OOD detection accuracy measured in AUC-ROC and AUC-PR (the higher, the better). The legend for each loss function is followed in parentheses with the selection criteria used for choosing hyperparameters. 
\edit{The results show the average and standard deviations.} 
Overall, the proposed weighted sum of $\uemdsq$ and UCE leads to the best prediction accuracy and steadily improving OOD detection performance when given more training samples. 
\edit{
Due to the distribution shift between the training and test environments, too much training data leads to degrading prediction accuracy for the other loss designs. 
In addition, compared to $\emdsq$-based losses, UCE is worse at capturing the cross-bin relationship among the discrete traction values, resulting in worse prediction accuracy and unstable OOD detection performance.}
}
\label{fig:loss_function_benchmark}
\vspace*{-0.2in}
\end{figure*}

\begin{table}[t]
\centering
\includegraphics[width=\linewidth, trim={0.cm -0.4cm 0.cm 0cm},clip]{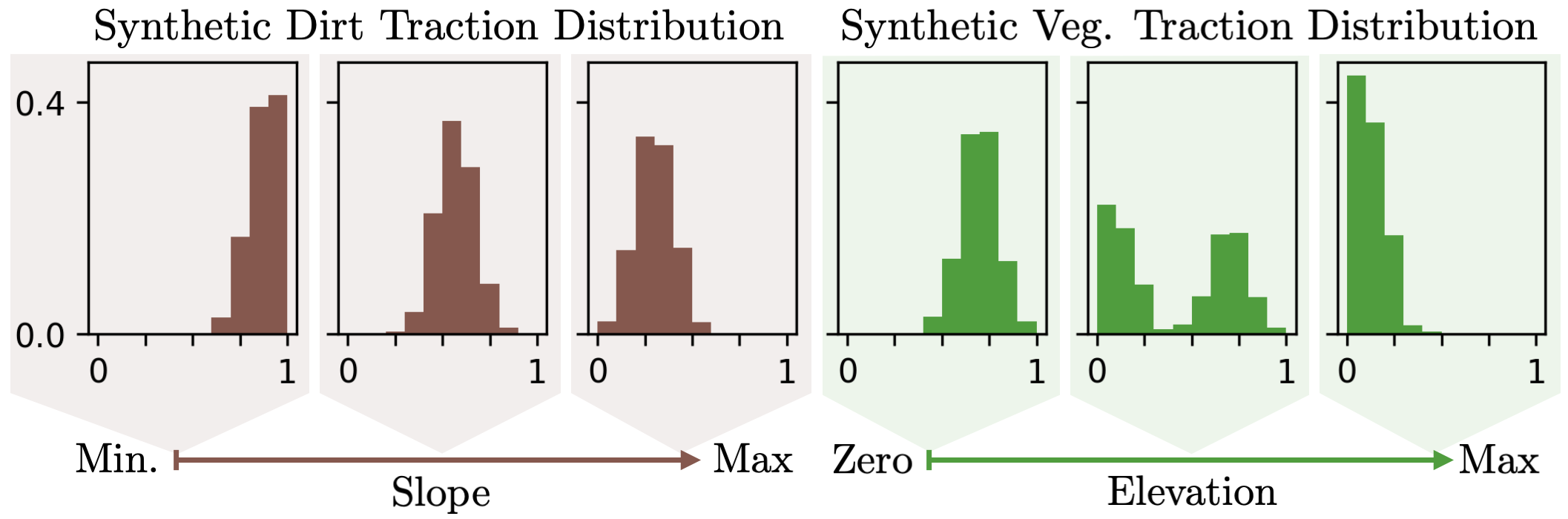}
\centering
\begin{tabular}{lllllll}
\hline
\multirow{2}{*}{Dataset Types} & \multicolumn{2}{l}{Elevation Range (m)} & \multicolumn{2}{l}{Max Slope} & \multirow{2}{*}{\begin{tabular}[c]{@{}l@{}}Veg.\\Ratio\end{tabular}} & \multirow{2}{*}{\begin{tabular}[c]{@{}l@{}}OOD\\Ratio\end{tabular}} \\ \cline{2-5}
& Dirt & Veg. & Dirt & Veg. & & \\ \hline
Train   & -0.2--0.0   & 0.3--0.7   & 0.3    & 0.4   & 0.2  & /    \\ %
Test   & -0.3--0.0  & 0.5--1.8 & 0.7   & 0.9  & 0.3  & /  \\ %
OOD (I)   & -0.5--0.1  & 0.4--1.8 & 0.7   & 1.0  & 0.3  & 0.5  \\ %
OOD (II)  & -0.6--2.0  & / & 0.9   & /  & 0.0  & 0.5  \\ \hline
\end{tabular}
\caption{Synthetic terrain dataset for benchmarking loss functions. The ground truth traction distributions for dirt are unimodal Gaussian distributions whose mean increases with terrain slope that indicates roughness of the terrain. Traction distributions for vegetation  are based on elevation, where the traction is bi-modal for intermediate elevation and uni-modal at the minimum and maximum elevations.
Note that OOD dataset (I) consists of mixed terrain types, but OOD (II) contains no vegetation to ensure that learned models do not rely solely on semantics for OOD detection.
}\label{tab:terrain_properties_loss_function_benchmark}
\vspace*{-0.2in}
\end{table}

The synthetic dataset contains randomly generated 3D terrain with ground truth traction distributions generated based on geometric properties (elevation and slope) and semantic types (dirt and vegetation); details are available in Table~\ref{tab:terrain_properties_loss_function_benchmark}. 
\edit{Note that terrain slopes are only used for generating the ground truth traction distributions, but are not used as inputs to the NN.}
For simplicity, we use the same traction distribution for both linear and angular components, and dependencies only exist between dirt and terrain slope, and vegetation and terrain elevation. While more complex traction distributions can be designed, our dataset is sufficient for supporting our contributions. 

In total, there are 5 training, 20 test, and 40 OOD environments that are 30 meters in width and height, 0.5 meters in resolution, as well as different elevations, slopes and vegetation ratios, where the training dataset is intentionally small in order to examine the generalization of learned models. Every training environment is split into equal parts for training and cross validation respectively. The synthetic environments are selectively visualized in Fig.~\ref{fig:synthetic_datasets}. To simulate real-world data collection, traction samples are only obtained along a circular path.
Moreover, we consider the impact of increasing the number of samples by multiplying the base measurement counts by factors $10^{k}$ where $k\in\{0,\dots,4\}$. 
\edit{For the training environments, the traction samples are accumulated in traversed terrain cells via histograms to obtain the empirical traction distributions, and the measurement counts are also stored in order to weight the training loss to discount the rarely visited terrain.
In the test environments, we use the ground truth traction distributions to measure the prediction accuracy of trained models.
In the OOD environments, terrain with slope and elevation values unseen in the training dataset are considered OOD. The associated OOD masks are the ground truth used to benchmark OOD detection performance. An example of the OOD mask is visualzied in Fig.~\ref{fig:synthetic_datasets}(c).}

\subsection{Model Training}\label{sec:model_training}
We use the same network architecture for all the loss functions, where the traction predictor
\edit{consists of a shared encoder (convolutional layers followed by fully connected layers) to process the semantic and elevation patches of the terrain, and two fully connected decoder heads with soft-max outputs for predicting the linear and angular traction distributions.} The latent space features \edit{from the shared encoder} are passed to a radial flow~\cite{rezende2015variational} and we use a constant certainty budget that scales exponentially with the latent dimension for numerical purposes~\cite{natpn}. 
\edit{During training, we follow the two-step procedure outlined in~\cite{natpn}. First, we jointly train the traction distribution predictor and the flow network. After convergence, we freeze the traction predictor and only fine-tune the flow network. This strategy improves OOD detection accuracy. However, we observe no improvement by performing ``warm-up training" for the flow network prescribed by~\cite{natpn}.}

Hyperparameter sweeps are conducted over learning rates in $[1\mathrm{e}{-4},3\mathrm{e}{-4},1\mathrm{e}{-3}]$ for the Adam optimizer, and entropy weights in $[0, 1\mathrm{e}{-6}, 1\mathrm{e}{-5}]$ when $\uemdsq$ and UCE are used separately. For the weighted sum of $\uemdsq$ and UCE, we fix the UCE term and consider additional weights for $\uemdsq$ in $[0.1, 1, 10]$. For each combination of hyperparameters, we train the model with five random seeds and select the best model based on validation $\emdsq$ error averaged over the seeds because empirically, we have found that selecting models based on validation $\emdsq$ instead of Kullback-Leibler (KL) divergence leads to improved performance for \textit{all} models. To guarantee fairness for the state-of-the-art and not clutter the figures, we only present the results for models selected based on validation KL divergence for the UCE loss.

\subsection{Prediction Accuracy and OOD Detection Performance}\label{sec:prediction_accuracy_and_ood_detection}

\edit{
Variations of the proposed loss function~\eqref{eq:uce_uemd2_h} are compared in terms of prediction accuracy and OOD detection performance. The prediction accuracy is measured by $\emdsq$ and KL divergence by comparing the predicted and the ground truth traction distributions. The accuracy of OOD detection using the densities of latent features is measured by area under the receiver operating characteristic curve (AUC-ROC) and area under the precision-recall curve (AUC-PR) with respect to the ground truth OOD masks.
}
Note that AUC-ROC and AUC-PR are standard metrics for binary classification that are invariant to scale and offset. Intuitively, a score of 0.5 means the classifier is as good as random guesses, and a score of 1 indicates a perfect classifier.
To show the best performance achievable by the state-of-the-art with unlimited traction samples during training, we include models trained with UCE using the \textit{ground truth traction distributions} in the training environments. 
\edit{
The benchmark results are in Fig.~\ref{fig:loss_function_benchmark}, where we report the average values and the standard deviations over all map cells,  test environments and random seeds.
}

The main takeaway from the benchmark is that the models trained with the proposed weighted sum of $\uemdsq$ and UCE achieves the best prediction accuracy in \textit{both} $\emdsq$ and KL divergence. Furthermore, the weighted-sum objective \edit{leads to} more stable improvements in test performance for both prediction accuracy and OOD detection as training samples become more abundant. 
\edit{
Interestingly, too many training samples lead to degrading prediction accuracy achieved by the other loss designs at test time. Because we do not observe worsening accuracy on validation dataset, the degrading test performance can be attributed to the distribution shift between the training and test environments as shown in Table.~\ref{tab:terrain_properties_loss_function_benchmark}.}
Notably, compared to $\emdsq$-based losses, UCE does not capture the cross-bin relationship of the traction distribution, which leads to worse regularized latent space that causes unstable OOD detection performance (even for UCE trained with ground truth traction distributions in the training environments).

\subsection{Ablation Study for $\uemdsq$ and Joint Training}\label{sec:uemd2_ablation}
While the benefits of using uncertainty-aware loss and joint training have been established in~\cite{natpn} for UCE, we present a similar ablation study for $\uemdsq$ for completeness in Table.~\ref{tab:ablation_uemd2}.
We set the sample multiplier to 10 for simplicity, but similar conclusions can be drawn with more samples. The takeaway is that both joint training and uncertainty awareness are required to achieve improved accuracy in $\emdsq$ and OOD detection.
Despite these improvements, the results in Fig.~\ref{fig:loss_function_benchmark} show that both $\uemdsq$ and UCE are required to achieve more consistent and steadily improving performance in prediction accuracy and OOD detection performance.

\begin{table}[h]
\centering
\caption{Ablation study for $\uemdsq$ and joint training. \edit{Results shown are the mean and standard deviation values obtained across random seeds.}}
\label{tab:ablation_uemd2}
\begin{tabular}{llll}
\toprule
Loss  & Test $\emdsq$ $\downarrow$ & AUC-ROC $\uparrow$  & AUC-PR $\uparrow$ \\ \midrule
\rowcolor[gray]{.9}
$\uemdsq$ (Joint) & $\mathbf{0.204 \pm 0.01}$ & $\mathbf{0.834 \pm 0.07}$ & $\mathbf{0.840 \pm 0.05}$  \\
$\emdsq$ (Joint)  & $0.236 \pm 0.02$  & $0.802 \pm 0.04$ & $0.830 \pm 0.03$ \\
$\emdsq$ (Disjoint)  & $0.228 \pm 0.03 $ & $0.665 \pm 0.14$  & $0.770 \pm 0.07$ \\ \bottomrule
\end{tabular}
\vspace*{-0.2in}
\end{table}

\section{Evaluation of Risk-Aware Planners}\label{sec:benchmark_planners}

\begin{figure}[t]
     \centering
    \includegraphics[width=\linewidth, trim={0cm 0cm 0cm 0cm},clip]{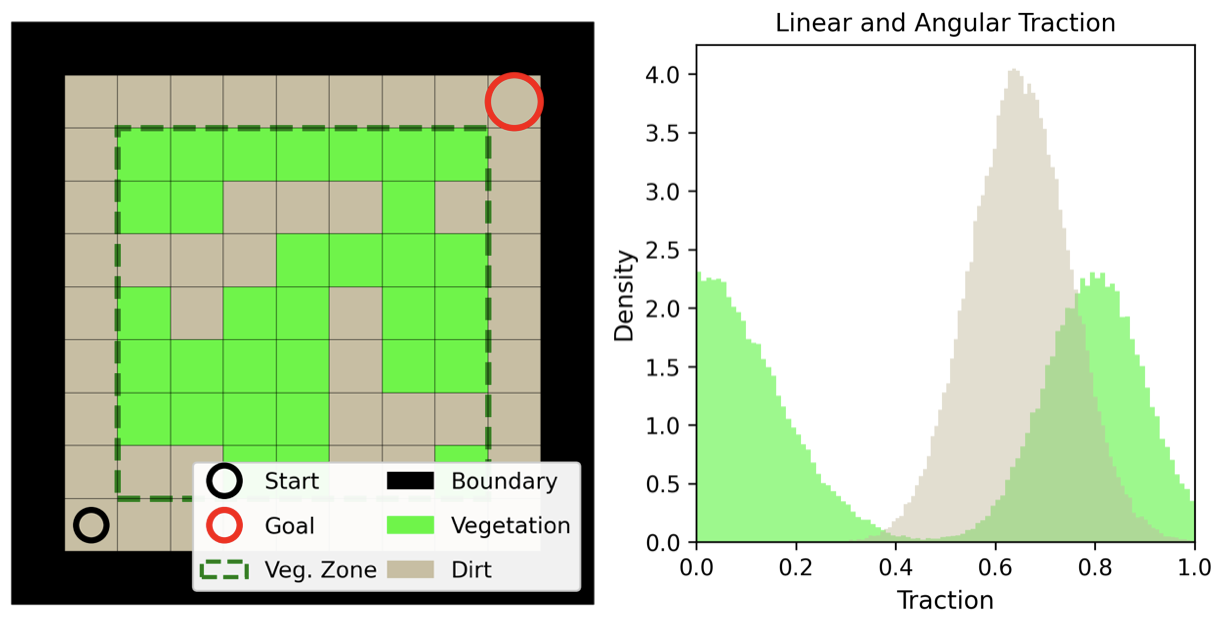}
        \caption{The simulation environment where a robot has to move from start to goal as fast as possible within the bounded arena. Linear and angular traction parameters share the same distribution for simplicity. Vegetation terrain patches are randomly sampled at the center in the vegetation zone.}
        \label{fig:sim_benchmark_setup}
  \vspace*{-0.2in}
\end{figure}

Using simulated 2D semantic environments \edit{whose terrain traction has high aleatoric uncertainty}, we show that the proposed \cvardyn{} outperforms existing approaches~\cite{Gasparino2022wayfast, cai2022risk} that assume the nominal traction or the expected traction, while achieving \edit{competitive performance compared to} \cvarcost{}~\cite{wang2021adaptive}. Moreover, we discuss the advantages and limitations of \edit{\cvardyn{} and \cvarcost{}} compared to the approach that assumes nominal traction while penalizing trajectories moving through terrain with high aleatoric uncertainty. 
For simplicity, we consider a grid world where dirt and vegetation cells have known traction distributions, as shown in Fig.~\ref{fig:sim_benchmark_setup}. Vegetation cells are randomly spawned with increasing probabilities at the center of the arena, and a robot may \edit{get stuck} due to vegetation's bi-modal traction distribution. \edit{The mission is successful if the robot reaches the goal without encountering zero-traction regions, colliding with obstacles, or getting stuck in local minima (e.g., when the robot does not move or just repeat circular trajectories without progressing to the goal).}

\subsection{Planner Implementation}\label{sec:mppi_planner_implementation_sim}
We adopt MPPI~\cite[Algorithm~2]{williams2017information} because it is derivative-free and parallelizable on GPU. The planners run in a receding horizon fashion with 100 timesteps at 0.1~s intervals. The maximum linear and angular speeds are 3~m/s and $\pi$~rad/s, and the noise standard deviations for the control signals are 2~m/s and 2~rad/s. The number of control rollouts is 1024, and the number of sampled traction maps is 1024 (only for \cvarcost{}). We use PMFs with 20 uniform bins to approximate the traction distribution.
A computer with Intel Core i9 CPU and Nvidia GeForce RTX 3070 GPU is used for the simulations, where the majority of the computation happens on the GPU. The \cvarcost{} planner is the most expensive to compute, but it is able to re-plan at 15~Hz while sampling new control actions and maps with dimensions of $200\times 200$. Planners that do not sample traction maps can be executed at over 50~Hz.

\subsection{Navigation Performance}\label{sec:planner_nav_performance_sim_2d}

\begin{figure*}[t]
	\centering
	\includegraphics[width=\linewidth, trim={0cm 0cm 0cm 0cm},clip]{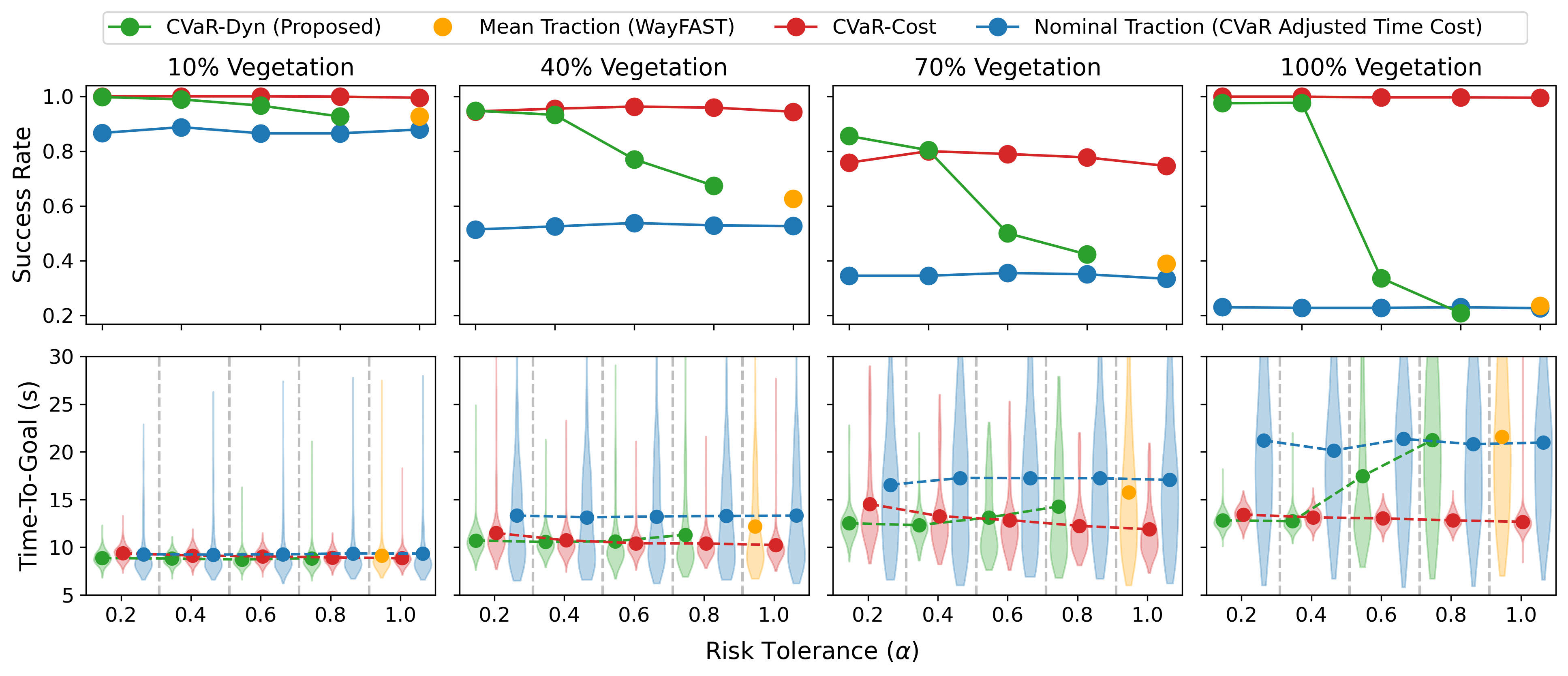}
 \caption{
 \edit{%
Success rates and time-to-goal achieved by the proposed \cvardyn{}, \cvarcost{}~\cite{wang2021adaptive}, WayFAST~\cite{Gasparino2022wayfast} that uses the expected traction or the method that assumes nominal traction~\cite{cai2022risk} (i.e., no slip). Note that a mission is successful if the robot reaches the goal.
We show the distributions of time-to-goal and their average values.
Overall, when the risk tolerance is sufficiently low (e.g., $\alpha=0.4$), \cvardyn{} achieves similar or better success rate and time-to-goal compared to the \cvarcost{} planner and outperforms both WayFAST and the method that assumes nominal traction. 
    }%
}\label{fig:sim_benchmark} 
 \vspace*{-0.2in}
\end{figure*}

\edit{We compare the proposed \cvardyn{} against \cvarcost{}~\cite{wang2021adaptive}, WayFAST~\cite{Gasparino2022wayfast} that uses the expected traction, and the technique in~\cite{cai2022risk} that assumes the nominal traction while adjusting the time cost with the CVaR of linear traction. Note that WayFAST is a vision-based navigation approach that predicts the expected terrain traction from images, but our analysis only focuses on the use of the expected traction values and its impact on the navigation performance.  We vary the conservativeness of all the methods (other than WayFAST) by changing the quantile $\alpha\in(0, 1]$ for computing the CVaR.}
Overall, we sample $40$ different semantic maps and $5$ random realizations of traction parameters for every semantic map. The traction parameters are drawn before starting each trial and remain fixed. 
\edit{
The benchmark results can be found in Fig.~\ref{fig:sim_benchmark}. The takeaway is that the proposed \cvardyn{} achieves better or similar success rate and time-to-goal when compared to \cvarcost{} if the risk tolerance $\alpha$ is sufficiently low. In addition, both \cvardyn{} and \cvarcost{} outperform the other methods that use the nominal or the expected traction values.}

\begin{figure}[t]
     \centering
    \includegraphics[width=\linewidth, trim={0cm 0cm 0cm 0cm},clip]{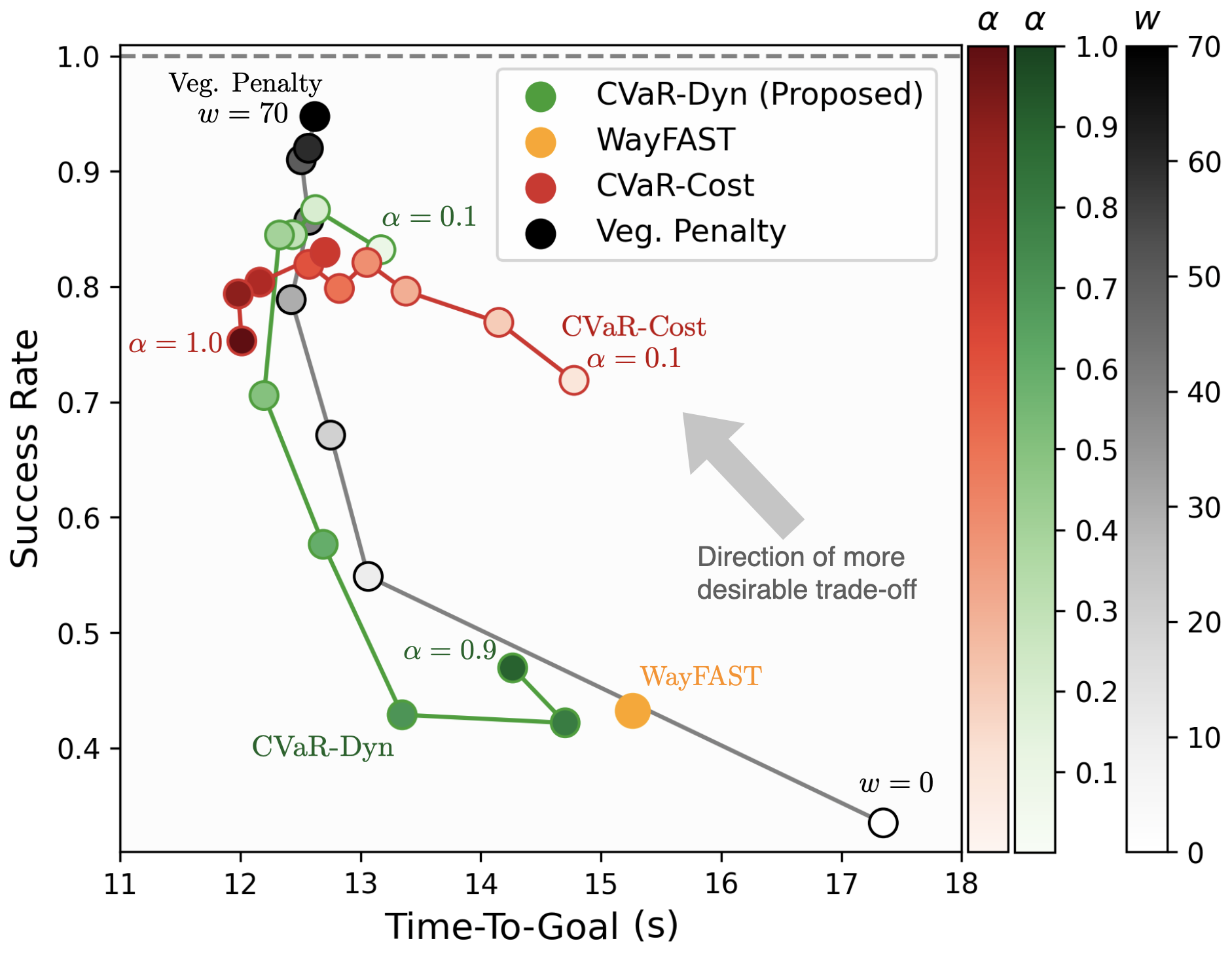}
\caption{
\edit{ %
Trade-offs between success rate and time-to-goal in the most challenging scenario of $70\%$ vegetation, where success is achieved if goal is reached. \cvardyn{} and \cvarcost{} both achieve better trade-offs than WayFAST by being in the upper left of the figure. When success rate is below 0.9, \cvardyn{} and \cvarcost{} achieve better trade-offs than the method that assumes nominal traction while imposing auxiliary penalty $w>0$ for states entering vegetation terrain. However, the success rates of \cvardyn{} and \cvarcost{} plateau and eventually degrade as $\alpha$ decreases, because the planners become more risk-averse and susceptible to local minima. 
} %
}\label{fig:veg_penalty}
  \vspace*{-0.2in}
\end{figure}

To compare CVaR-based methods against \edit{another baseline that plans with the nominal traction while imposing} auxiliary penalties for vegetation terrain with high aleatoric uncertainty, we focus on the most challenging setting with 70\% vegetation, where it is easy to get stuck in local minima. 
\edit{
To adjust risk tolerance, we consider $\alpha\in(0, 1]$ for CVaR-based methods and vegetation penalty $w\geq0$ for the planner that assumes nominal traction. The benchmark result in Fig.~\ref{fig:veg_penalty} shows the trade-offs between success rate and time-to-goal achieved by different methods (WayFAST included as a special case of \cvardyn{} when $\alpha=1$). 
Overall, all methods except WayFAST can be tuned to improve success rate and time-to-goal. Assigning high vegetation penalties leads to the best success rate, because we observe that the robot always avoids the vegetation terrain. On the other hand, \cvardyn{} and \cvarcost{} can achieve better time-to-goal at a lower success rate, which may be desirable for more risk-tolerant and time-critical missions. As $\alpha$ decreases further, \cvardyn{}'s performance first plateaus and then worsens, because the state rollouts become too short when using the worst-case expected traction, making \cvardyn{} susceptible to local minima. While \cvarcost{} also experiences worsening performance as $\alpha$ decreases, its conservativeness is caused by the greater difficulty of estimating CVaR of objective.
Overall, none of methods completely dominate the others. Therefore, when domain knowledge is available, auxiliary penalties for undesirable terrain can be used together with CVaR-based methods to improve performance (see Sec.~\ref{sec:exp_bag_sim}). 
While the simulation shows comparable performance achieved by \cvardyn{}, \cvarcost{} and the baseline, we show that \cvardyn{} achieves the best performance in practice (see Sec.~\ref{sec:results:hardware_experiments}). 
}

\section{Optimizing for $\emdsq$ Improves Navigation}\label{sec:emd_and_nav_performance}

\begin{figure}[t]
\centering
\includegraphics[width=\linewidth, trim={0.cm 0.cm 0.cm 0.cm},clip]{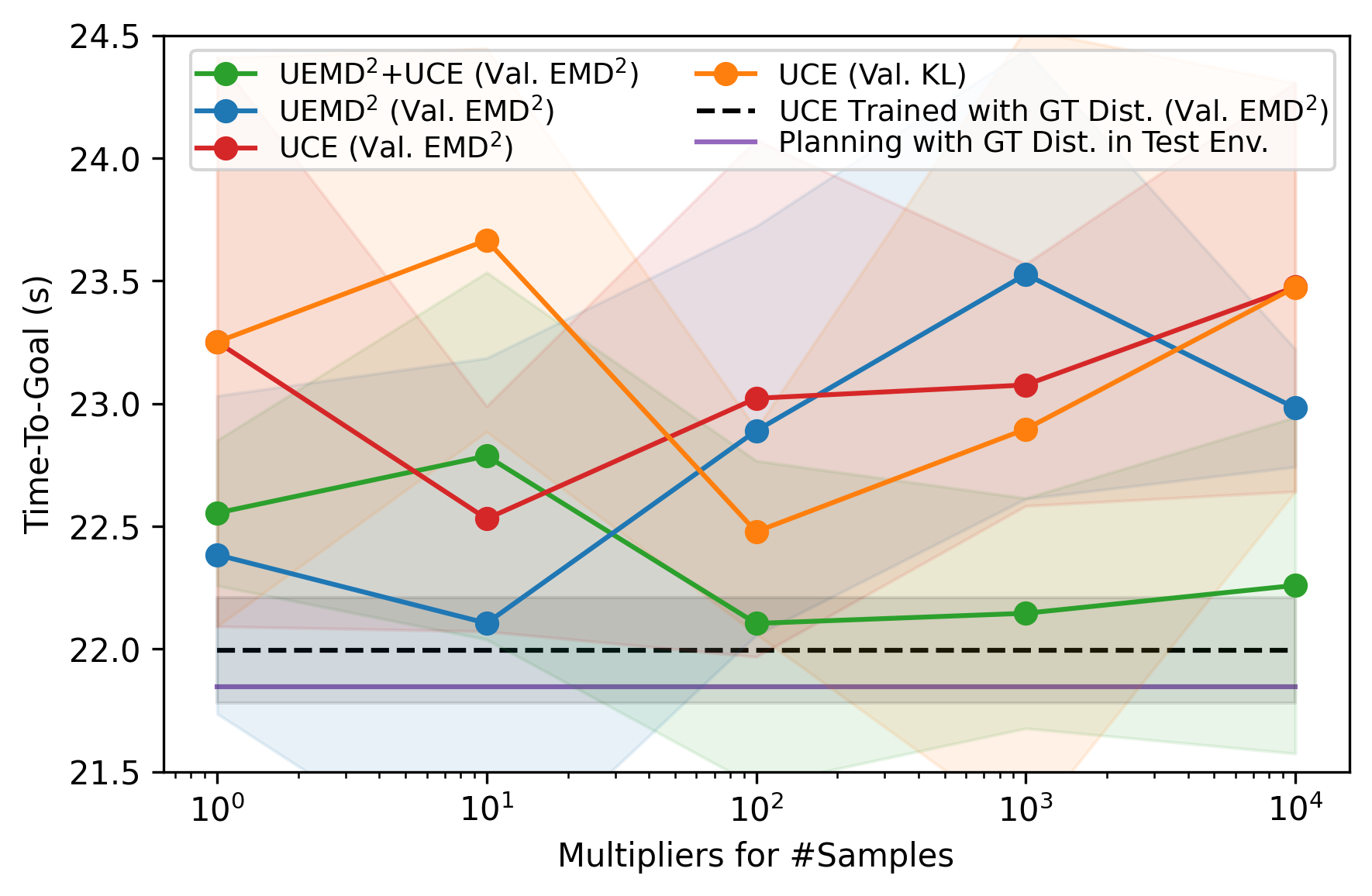}
\caption{Navigation performance using learned traction models trained with different loss designs in the test environments shown in Fig.~\ref{fig:synthetic_datasets}. 
\edit{The results show the average values and the standard deviations over all terrain cells, test environments, and random seeds.} 
Note that the navigation performance of the proposed hybrid loss approaches the best possible navigation performance using the ground truth (GT) traction models in the test environments and the best possible navigation performance of the state-of-the-art UCE loss trained with the GT traction distributions in the training environments.}
\label{fig:emd2_nav_performance}
\vspace*{-0.2in}
\end{figure}

To support the key argument that $\emdsq$ is a better metric than KL divergence for measuring the quality of learned traction distributions for navigation, we evaluate the navigation performances when using models trained with different losses presented in Sec.~\ref{sec:benchmark_traversability_learning_pipeline}. The models are deployed in the same test environments visualized in Fig.~\ref{fig:synthetic_datasets}, where each map is 30~m in width and height and the start and goal positions are at the opposite diagonal corners. To not clutter the results, we only focus on the proposed \cvardyn{} planner with $\alpha=0.4$ and the same MPPI setup in Sec.~\ref{sec:mppi_planner_implementation_sim}, but similar trend can be observed with different choices of $\alpha$.
Consistent with the loss benchmark in Sec.~\ref{sec:benchmark_traversability_learning_pipeline}, each loss is trained with 5 random seeds and 5 levels of data abundance. For each of the 20 test maps, we consider 5 randomly sampled traction maps and run the mission 3 times. The final results averaged over training seeds can be found in Fig.~\ref{fig:emd2_nav_performance}, where all trials are successful, so the success rate is omitted. 

Importantly, when the amount of \edit{training} data is low, $\uemdsq$ outperforms UCE in time-to-goal even though $\uemdsq$ leads to worse KL error than UCE loss as shown in Fig.~\ref{fig:loss_function_benchmark}. This validates our intuition that $\emdsq$ captures the cross-bin information of discretized traction values better, which facilitates the learning of traction distribution in low-data regime and leads to better navigation performance. 
\edit{
As more training data becomes available, the proposed weighted sum of $\uemdsq$ and UCE outperforms the other loss designs. 
Due to the distribution shift between the training and test environments as discussed in Sec.~\ref{sec:prediction_accuracy_and_ood_detection}, the traction prediction accuracy may degrade when given too much training data, resulting in degrading navigation performance observed in Fig.~\ref{fig:emd2_nav_performance}. However, the proposed hybrid loss is less susceptible to the distribution shift, thus sustaining the navigation performance better than the other methods.
}
Furthermore, the navigation performance of the proposed hybrid loss approaches the best possible performance of the state-of-the-art UCE loss when trained on ground truth traction distributions in the training environments, indicating good generalization of our approach using only \edit{the limited data collected along circular paths in the training environments}. For reference, the figure also provides the lower bound for the \edit{time-to-goal} based on the ground truth traction models in the test environments.

\section{Benefits of Avoiding OOD Terrain} %
\label{sec:exp_bag_sim}

\begin{figure}[t]
 \centering
\includegraphics[width=\linewidth, trim={0cm 0.3cm 0.cm 0cm},clip]{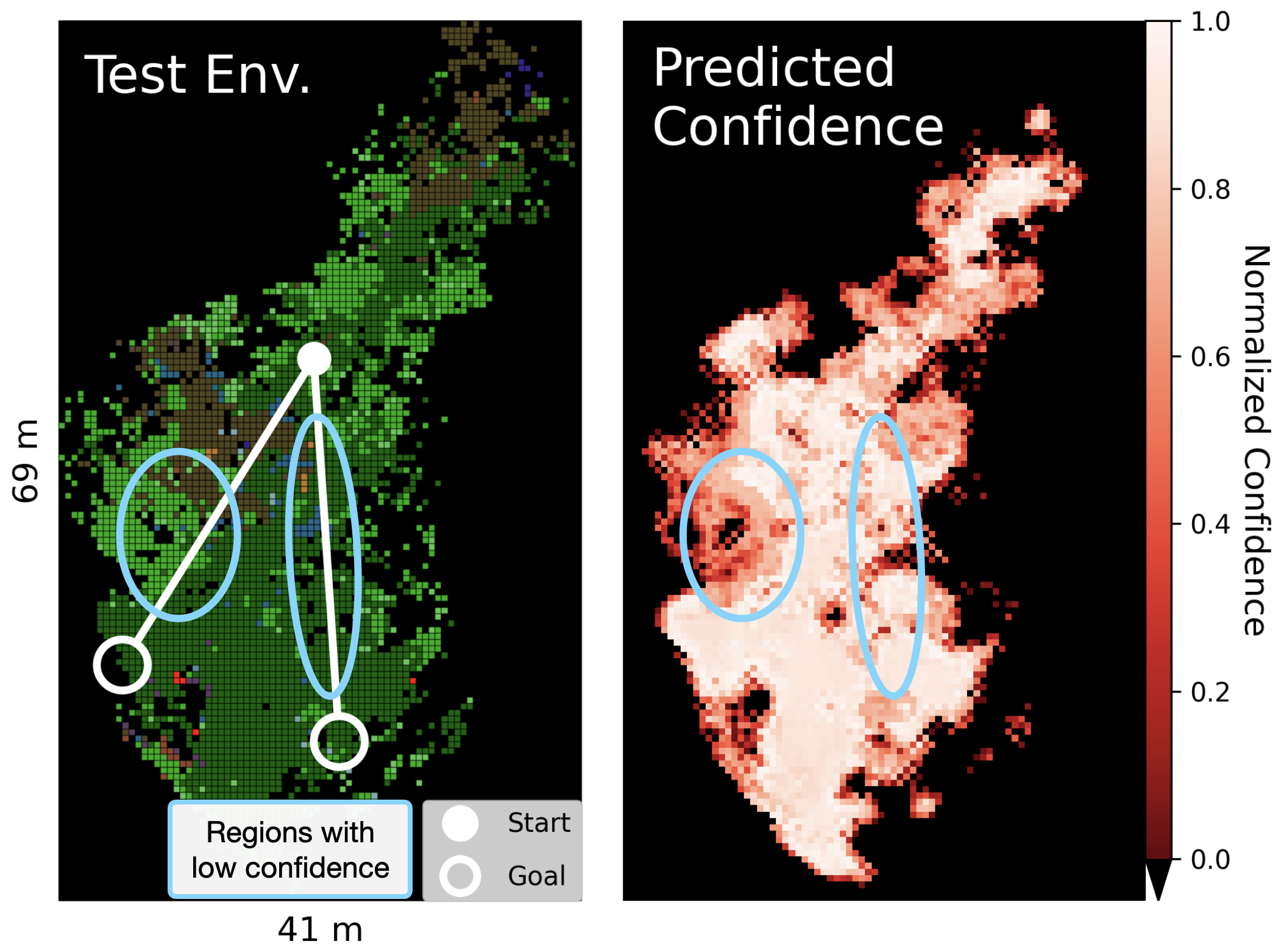}
\caption{(Left) In the test environment, the simulated robot has to reach two goals selected to highlight the danger of using unreliable network predictions. (Right) The latent density-based confidence score~\eqref{eq:latent_conf} indicates the amount of epistemic uncertainty for the predicted traction distribution, where unknown terrain and known terrain with negative scores are shown in black. Note that the brown semantic region (mulch) at the top has confidence below zero due to the presence of unknown cells, in contrast to the brown semantic region to the left with much fewer unknown cells.}
\label{fig:test_env}
\vspace*{0.10in}
\centering
\includegraphics[width=\linewidth, trim={0cm 0.2cm 0cm 0cm},clip]{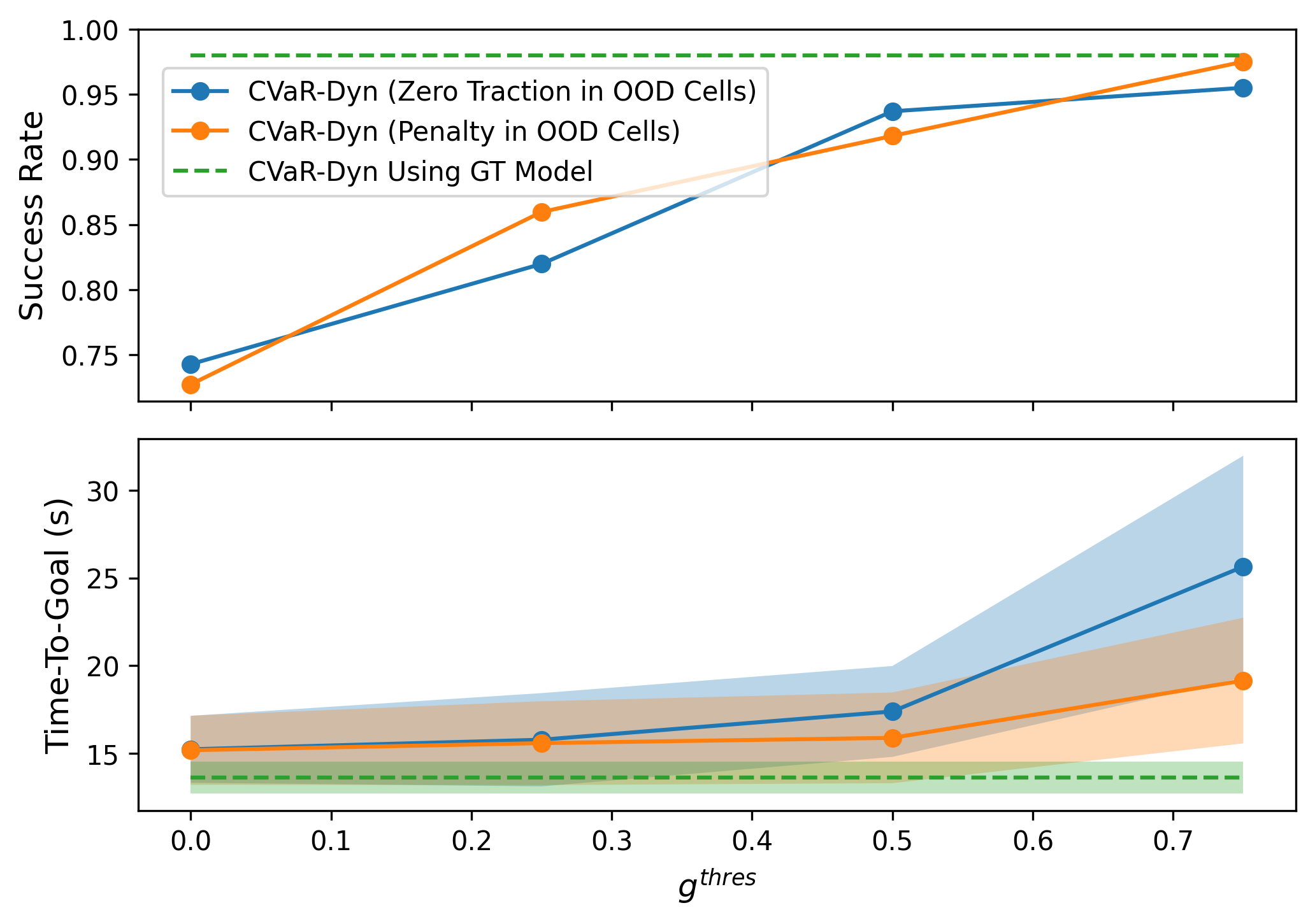}
\caption{\edit{Navigation success rate improves by avoiding OOD terrain. Note that the shaded areas indicate standard deviations. The OOD terrain is handled by either assigning zero traction (blue) or imposing penalties (orange). The performance of the planner that uses the ground truth (GT) traction is included to show the best possible performance. Overall, higher $g^{\text{thres}}$ improves the success rate at the cost of worse time-to-goal. However, auxiliary penalties for OOD terrain make it easier for the planner to find solutions that lead to the goal. Notably, the average success rate when $g^{\text{thres}}=0.75$ approaches $1$, indicating that the learned traction model generalizes well to terrain with high confidence values (low epistemic uncertainty) in the test environment.
}}
\label{fig:conf_score_benchmark} 
\vspace*{-0.2in}
\end{figure}

\begin{figure*}[!t]
 \centering
\includegraphics[width=\linewidth, trim={0cm 0cm 0cm 0cm},clip]{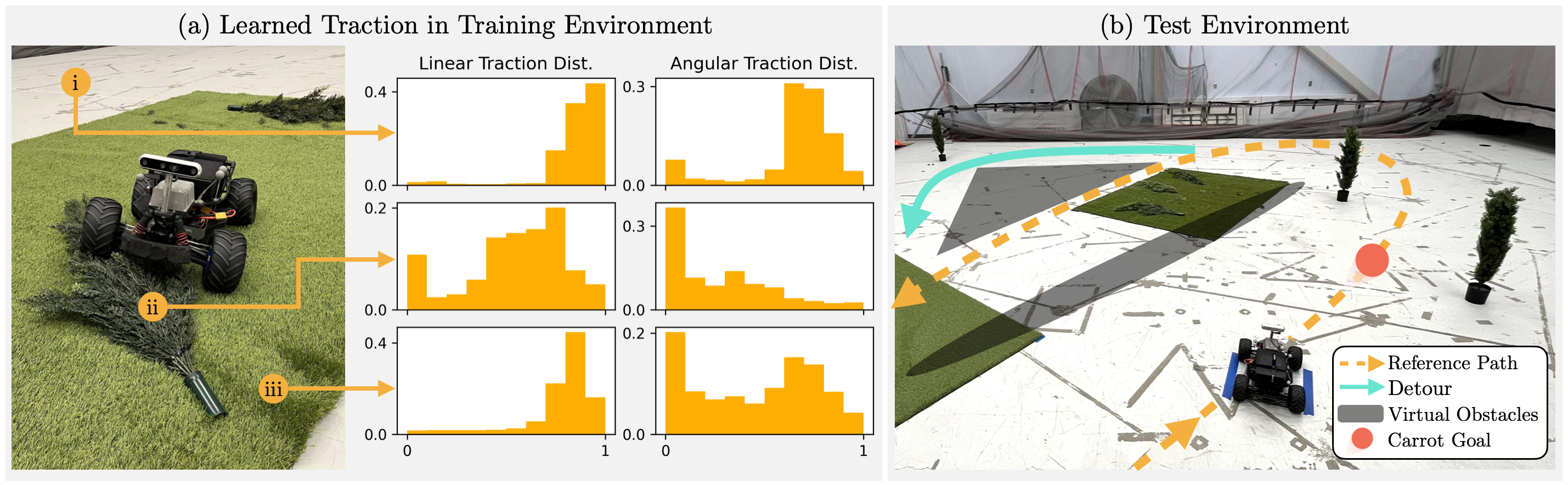}
\caption{The training and test environments used for the indoor racing experiments. 
(a)~The training environment consisted of a single turf with two fallen trees for simulating bushes. Learned linear and angular traction distributions are visualized for selected regions with (i)~hard floor, (ii)~fallen tree, and (iii)~turf. Note that the bi-modality of traction distribution over the vegetation could cause the robot to slow down significantly.
(b)~The test environment contained two turfs, three fallen trees, three standing trees, and virtual obstacles.
The robot was tasked to drive two laps following a carrot goal along the reference path while deciding between a detour without vegetation and a shorter path with vegetation.}
\label{fig:indoor_exp_setup_highlevel}
\vspace*{-0.2in}
\end{figure*}

We demonstrate the benefit of the proposed density-based confidence score~\eqref{eq:latent_conf} for detecting terrain with high epistemic uncertainty. To simulate training and test environments, we leverage the data collected in two distinct forests using Clearpath Husky, where the first one (visualized in Fig.~\ref{fig:data_collection}) is used for training, and the second one (whose semantic top-down view is shown in Fig.~\ref{fig:test_env}) is used as the test environment. 
\edit{The environment models were built using semantic octomaps~\cite{asgharivaskasi2021active} that fused lidar points and segmented RGB images based on the 24 semantic categories in the RUGD dataset~\cite{RUGD2019IROS}.}
The traction values will be drawn from the test environment's empirical traction distributions learned by a separate NN as the proxy ground truth. 
\edit{We use the proposed \cvardyn{} with a low risk tolerance $\alpha=0.2$ to handle the noisy terrain traction. }
Two specific start-goal pairs have been selected to highlight the most challenging parts of the test environment with novel features. Each start-goal pair is repeated 10 times for each selected confidence threshold $g^{\text{thres}}$. We investigate two ways to prevent the planner from entering terrain that is \edit{classified as OOD by either assigning zero traction, or adding large penalties. The mission is deemed successful if the goal is reached.} 

As shown in Fig.~\ref{fig:conf_score_benchmark}, the success rate improves by up to $30\%$ \edit{as the confidence threshold} $g^{\text{thres}}$ increases, because the robot avoids regions with unreliable traction predictions. Interestingly, using \cvardyn{} with soft penalties for OOD terrain leads to better time-to-goal while retaining a similar success rate, because the auxiliary costs for OOD terrain make it easier for the planner to find trajectories that avoid the OOD terrain. Therefore, it is advantageous to use \cvardyn{} with auxiliary costs when domain knowledge is available to achieve both a high success rate and fast navigation in practice.

\section{Hardware Experiments}\label{sec:results:hardware_experiments}
\edit{To evaluate the effectiveness and feasibility of EVORA (the overall framework for uncertainty-aware traversability learning and risk-aware planning)} in practice, we designed two experiment scenarios---an indoor race track scenario with fake vegetation using an RC car (Sec.~\ref{sec:results:indoor_exp_rc}) and a more challenging outdoor scenario using a legged robot (Sec.~\ref{sec:results:outdoor_exp_spot}). 
\edit{
For both scenarios, the robots used onboard sensors to map the environments online at test time, introducing more uncertainty due to motion blurs, lighting changes and incomplete maps.}
While both scenarios show that the proposed \cvardyn{} planner leads to the best navigation performance, the outdoor scenario also shows the benefits of avoiding OOD terrain. 
In practice, the control signals generated by MPPI are very noisy, so we plan in the derivative space of the nominal control~\cite{kim2022smooth} to generate smooth trajectories.

\subsection{Indoor Racing with an RC Car}\label{sec:results:indoor_exp_rc}

\begin{figure*}[t]
\centering
\includegraphics[width=\linewidth, trim={0cm 0cm 0cm 0cm},clip]{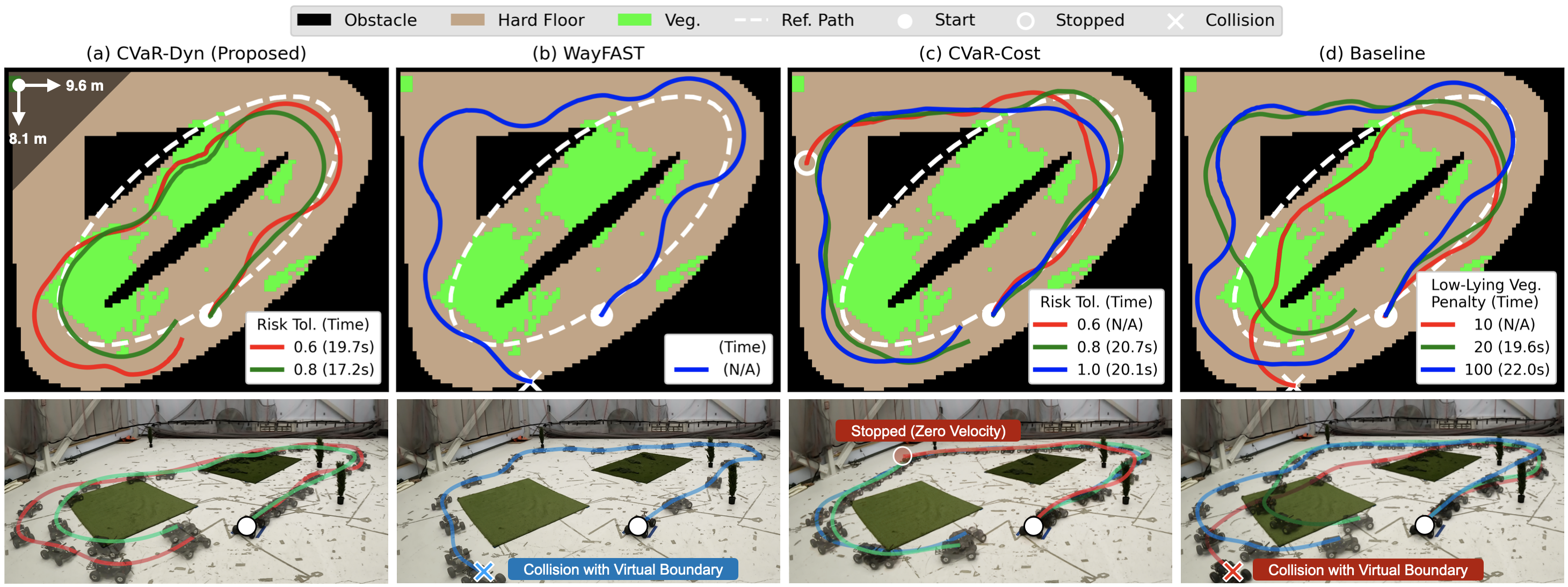}
\caption{
\edit{
Representative trials of the indoor experiments for highlighting the failure modes of the planners. The top-down semantic maps are shown in the top row and the time-lapse photos are shown in the bottom row. We only show the first lap out of the two laps for clarity. 
(a)~As $\alpha$ decreased, the proposed \cvardyn{} became more risk-averse and took wider turns in order to enter the shortcut.
(b)~WayFAST (\cvardyn{} with $\alpha=1$) did not account for the risk of under-steering, so it always turned too late for the shortcut.
(c)~\cvarcost{} consistently took the detour to avoid the vegetation terrain. As $\alpha$ decreased, the planner became more risk-averse and sometimes stopped near obstacles.
(d)~When the soft penalty was low, the baseline was more risk-tolerant and chose to take the shortcut, but the experienced traction differed significantly from the nominal traction, which caused more collisions. As the soft penalty increased, the planner became more conservative and took the detour, but planning with nominal traction led to significant under-steering that limited performance.
}
}
\label{fig:indoor_exp_nav_qualitative}
\vspace*{-0.2in}
\end{figure*}

\begin{figure}[t]
\centering
\includegraphics[width=\linewidth, trim={0.1cm 0.1cm 0.1cm 0.1cm},clip]{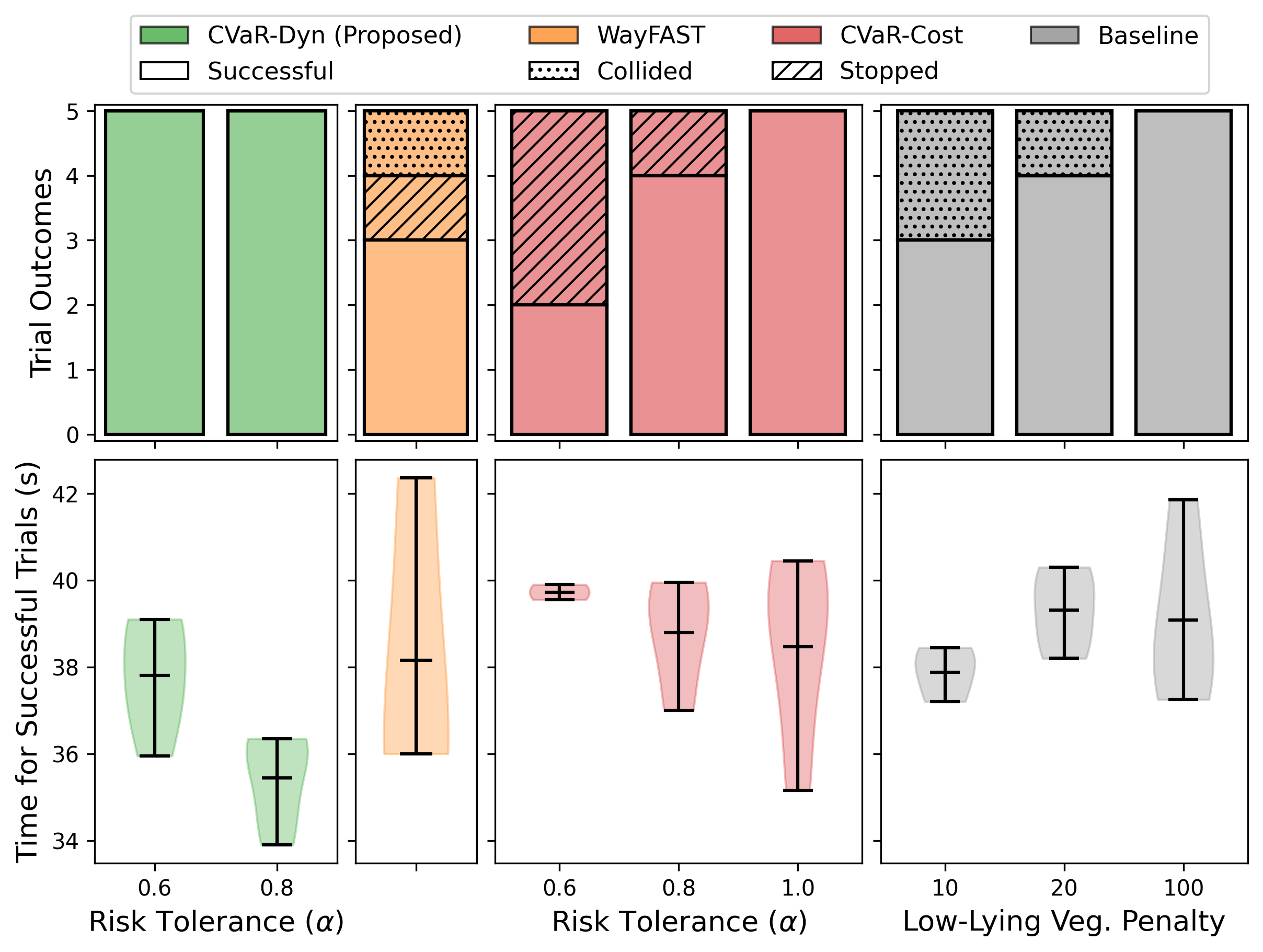}
\caption{
\edit{Outcomes and mission time for the indoor experiments over 5 trials. We show the distributions of mission time and the  maximum, average, and minimum values. The proposed \cvardyn{} with $\alpha=0.8$ achieved the best time-to-goal with 100\% success rate. As $\alpha$ reduces, \cvardyn{} and \cvarcost{} both led to worse time-to-goal. Notice that \cvarcost{} stopped near obstacles for many occasions when $\alpha<1$. In comparison, the baseline and WayFAST led to worse time-to-goal and a higher chance of collision.}
}
\label{fig:indoor_exp_nav_results}
\vspace*{-0.2in}
\end{figure}

\edit{
The goal of the indoor experiments is to show the performance benefits of the proposed planner for mitigating the risk due to aleatoric uncertainty in a controlled environment.
}

\subsubsection{\edit{Experiment Setup}}
An overview of the indoor setup is provided in Fig.~\ref{fig:indoor_exp_setup_highlevel}, which shows the 9.6~m by 8~m arena populated with turf and fake trees used to mimic outdoor vegetation. The 0.33~m by 0.25~m RC car was equipped with a RealSense D455 depth camera, an Intel Core i7 CPU, and a Nvidia RTX 2060 GPU. The robot ran onboard traction prediction, motion planning, and online elevation mapping with 0.1~m resolution, but Vicon was used for ground truth pose and velocity estimation. We identified vegetation by extracting the green image pixels instead of using a standalone NN semantic classifier in order to conserve GPU resources.
The bicycle model~\eqref{eq:bicycle} was used for this experiment, and the traction values were obtained by analyzing the commanded linear velocities, steering angles, and the ground truth velocities from Vicon. 

The traction model was trained based on 10~min of driving data with the proposed loss function~\eqref{eq:uce_uemd2_h}, where $\uemdsq$ and UCE were both weighted by~1 and the entropy term was weighted by $1\mathrm{e}{-5}$ based on empirical tuning. The learned traction distributions are visualized in Fig.~\ref{fig:indoor_exp_setup_highlevel}a to highlight multi-modality.
At deployment time, the robot ran 2 laps around the race track along the ellipsoidal reference path, while deciding between a shorter path covered with vegetation or a less risky detour, as shown in Fig.~\ref{fig:indoor_exp_setup_highlevel}b. We designed a moving goal region along the reference path, called the ``carrot goal'', that maintained a constant 75 degree offset from the robot's projected position on the ellipsoidal reference path. In addition to \cvarcost{} and the proposed 
\cvardyn{}, we considered an intelligent baseline that assumes nominal traction but assigns auxiliary penalties for low-lying vegetation between 5~cm and 15~cm that could cause unfavorable driving conditions. All methods avoided the trees via auxiliary penalties. All planners considered 1024 rollouts while planning at 20~Hz with 5~s look-ahead. Due to computational constraints, \cvarcost{} only considered 400 traction map samples. We set the maximum linear speed and steering angle to be 1.5~m/s and 30 degrees.

\subsubsection{\edit{Aleatoric Uncertainty Results}}
\edit{
The qualitative and quantitative results comparing planners' abilities to mitigate the risk due to aleatoric uncertainty are summarized in Fig.~\ref{fig:indoor_exp_nav_qualitative} and Fig.~\ref{fig:indoor_exp_nav_results}. We considered 3 risk tolerances $\alpha\in\{0.6, 0.8, 1\}$ for \cvardyn{}, \cvarcost{} and vegetation penalties $w\in\{10, 20, 100\}$ for the baseline that assumes nominal traction while penalizing states entering vegetation terrain. We present results for WayFAST~\cite{Gasparino2022wayfast} separately, but it is a special case of \cvardyn{} when $\alpha=1$. We repeated the race 5 times and each race consisted of 2 laps.
Overall, \cvardyn{} with $\alpha=0.8$ achieved the best time-to-goal and success rate. Qualitative visualizations in Fig.~\ref{fig:indoor_exp_nav_qualitative} show that the baseline and WayFAST both suffered from noisy real-world traction, causing wide turns. In comparison, \cvarcost{} and the proposed \cvardyn{} handled the noisy terrain traction better by producing smoother trajectories. Different from \cvardyn{}, the \cvarcost{} planner more frequently took the detour and sometimes got stuck in local minima near obstacles. 
}

\subsection{Outdoor Navigation with a Legged Robot}\label{sec:results:outdoor_exp_spot}

\begin{figure*}[t]
 \centering
\includegraphics[width=\linewidth, trim={0cm 0cm 0cm 0cm},clip]{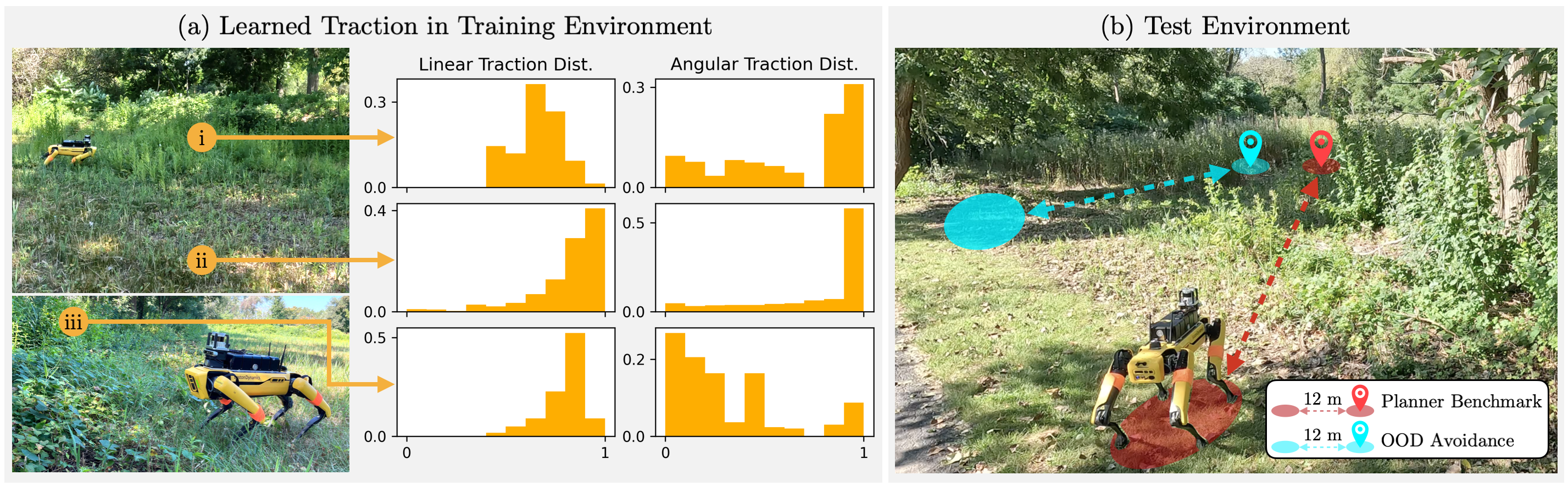}
\caption{The outdoor training and test environments with a legged robot.
(a)~The outdoor environment consisted of vegetation terrain with different heights and densities. Predicted linear and angular traction distributions are visualized for selected regions with (i) tall grass, (ii) short grass, and (iii) dense bushes. Unlike wheeled robots, a legged robot typically has good linear traction through vegetation, but angular traction may exhibit multi-modality due to the greater difficulty of turning.
(b)~Two start-goal pairs were used to benchmark the planners and analyze the benefits of avoiding OOD terrain.}
\label{fig:spot_exp_setup_highlevel}
\centering
\includegraphics[width=0.85\linewidth, trim={0cm 0.cm 0cm 0.cm},clip]{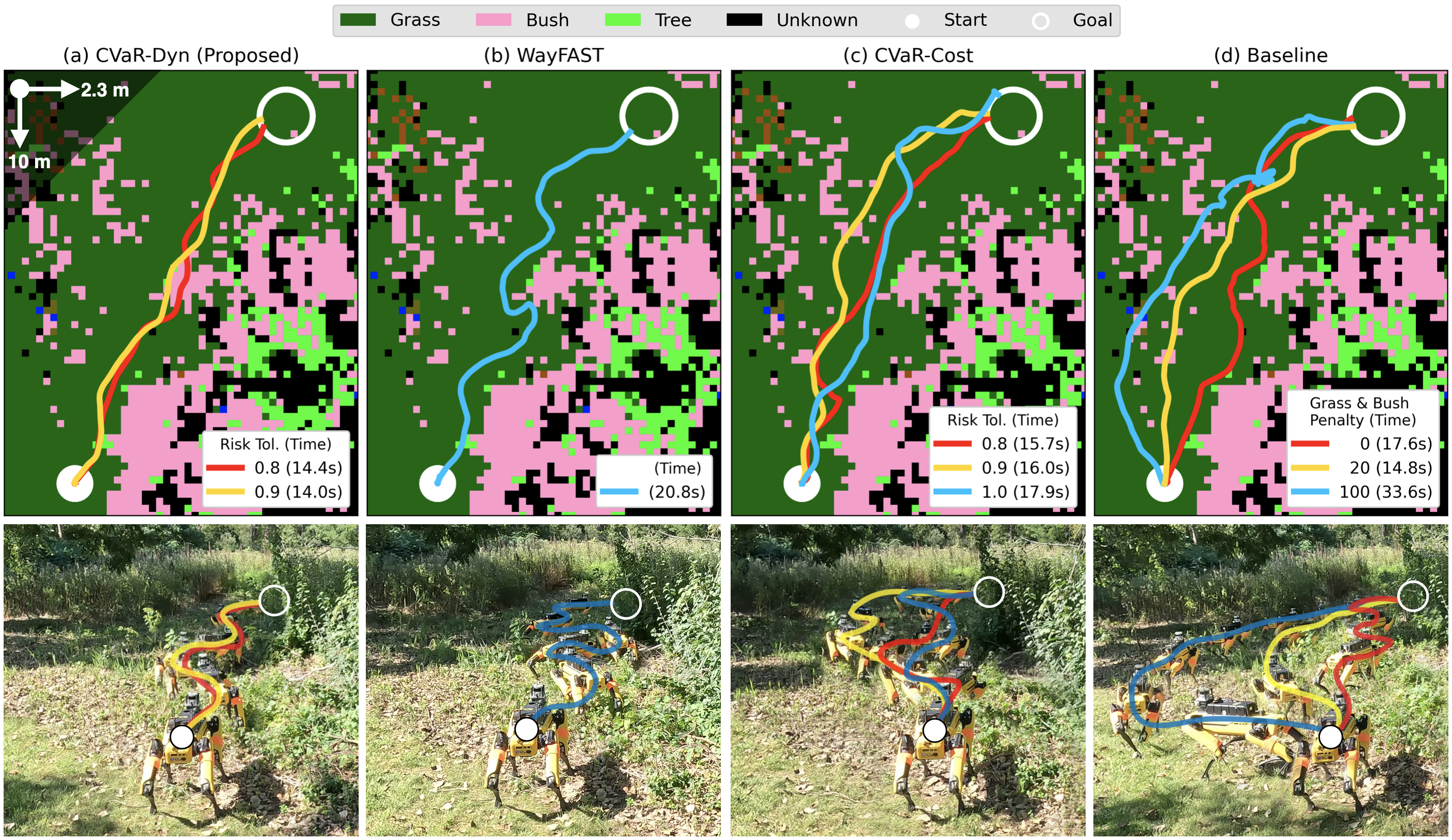}
\caption{
\edit{
Representative trials of the outdoor experiments. The top-down semantic maps are shown in the top row and the time-lapse photos are shown in the bottom row.
(a)~The proposed \cvardyn{} with $\alpha<1$ handled the noisy terrain traction well and produced less wavy trajectories compared to other methods. 
(b)~WayFAST (\cvardyn{} when $\alpha=1$) relied on the expected traction that provided a poor indication of the actual trajectory outcome, causing the constant correction in headings.
(c)~\cvarcost{} was more conservative compared to \cvardyn{} by staying further away from bushes and achieved longer time-to-goal.
(d)~The baseline assumed nominal traction that led to under-steering. As soft penalties increased, the robot became more averse to tall grass and bushes. As most of the test area was filled with grass or bush, the baseline with large soft penalties struggled to find feasible plans to goal in subsequent trials.
}
}\label{fig:spot_planner_qualitative}
\vspace*{-0.2in}
\end{figure*}

\begin{figure}[t]
\centering
\includegraphics[width=\linewidth, trim={0cm 0.cm 0cm 0.cm},clip]{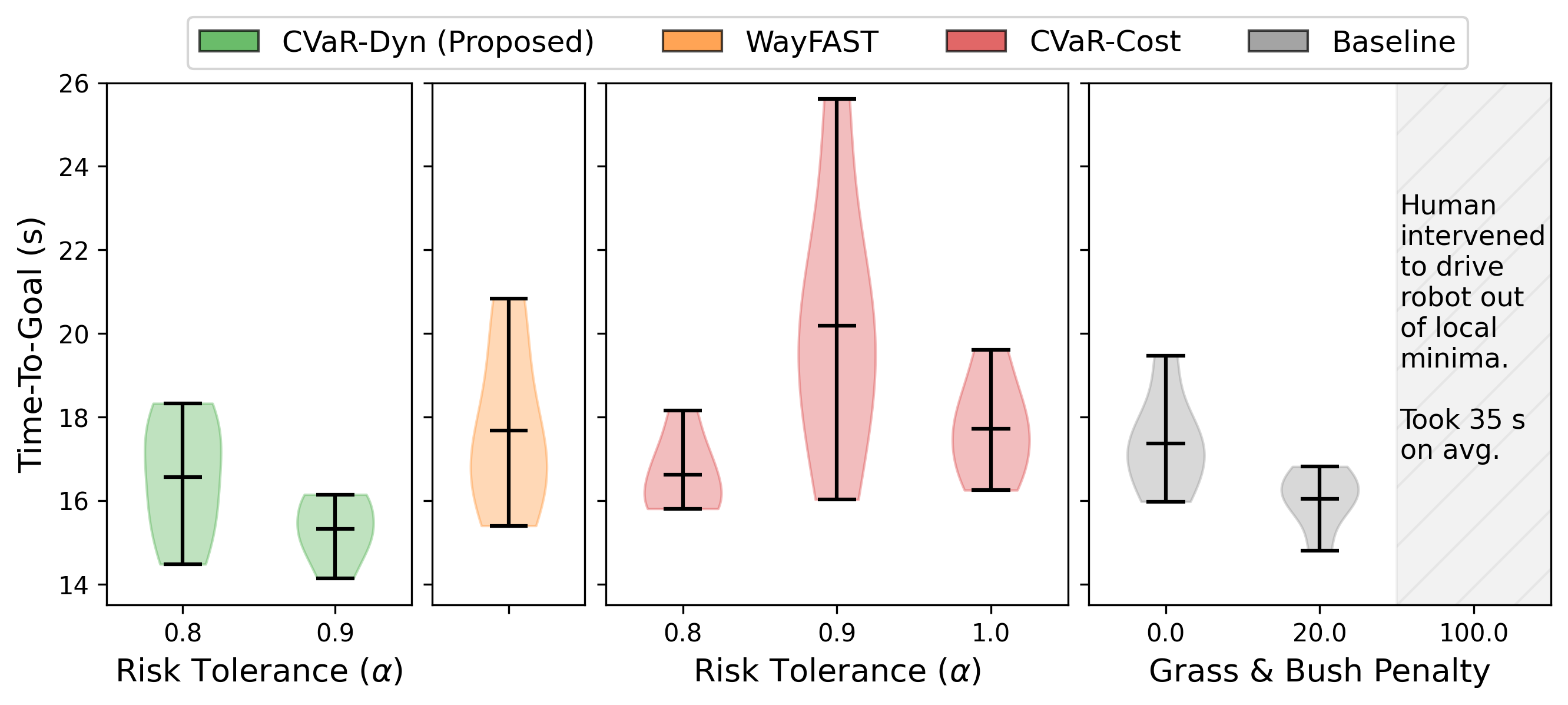}
\caption{\edit{Distributions of time-to-goal for the local planner benchmark with maximum, average, and minimum values. Each planner completed 3 round trips or 6 trials in total.} The proposed \cvardyn{} with $\alpha=0.9$ outperformed \cvarcost{} that required more computation, WayFAST (\cvardyn{} with $\alpha=1$) that planned with the expected traction, and the baseline that planned with the nominal traction and assigned soft penalties for grass and bushes.}
\label{fig:spot_planner_quantitative}
\vspace*{-0.2in}
\end{figure}

\begin{figure}[t]
\centering
\includegraphics[width=\linewidth, trim={0cm 0.cm 0cm 0.cm},clip]{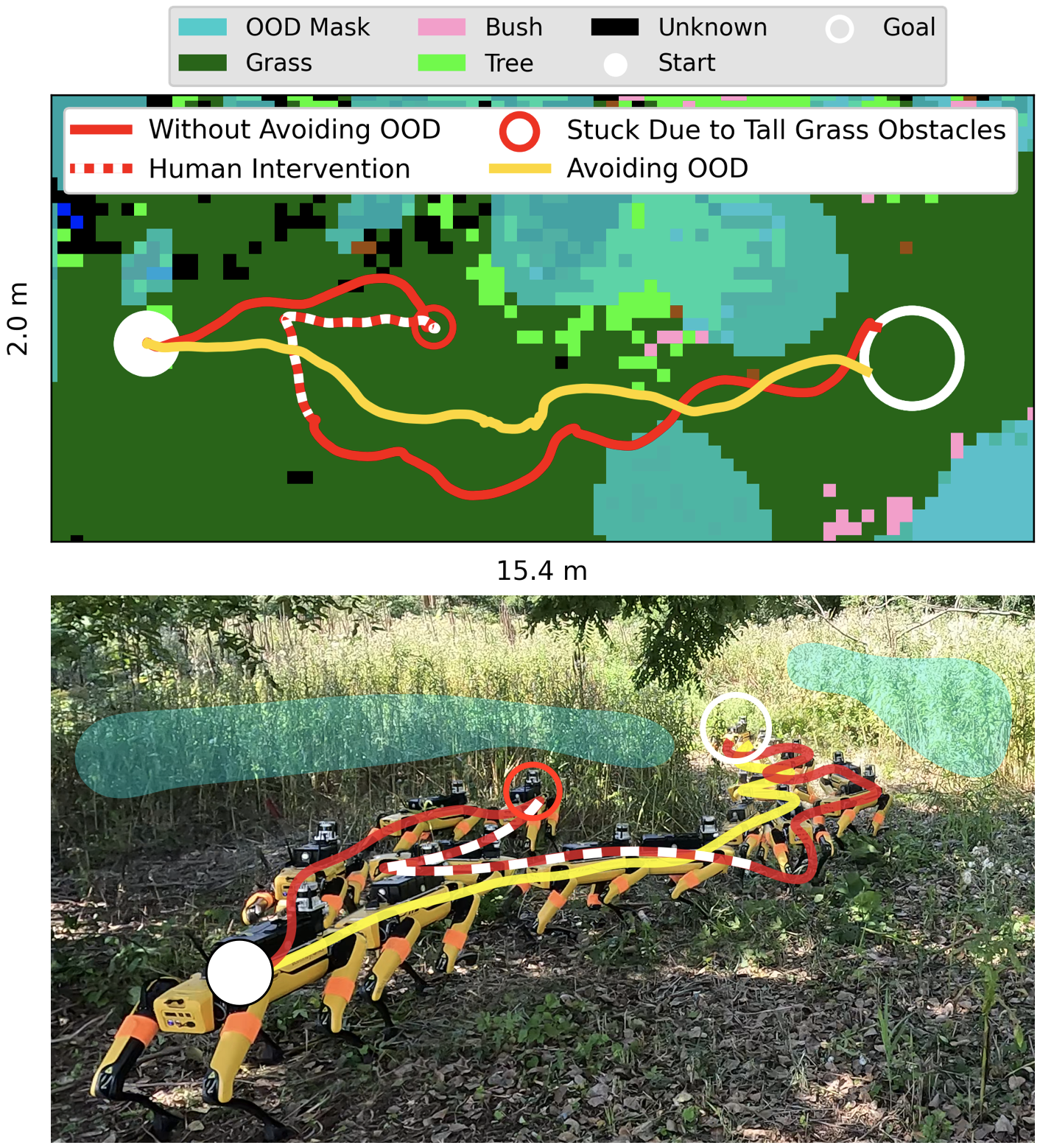}
\caption{
\edit{
Representative planner behaviors that show the benefits of avoiding OOD terrain, where the semantic top-down map and the time-lapse photo are shown on the top and bottom. Without OOD avoidance, the robot was more susceptible to local minima due to imperfect online map and noisy terrain traction, requiring human interventions to teleoperate the robot to a region with feasible plans to goal. In contrast, assigning auxiliary penalties for OOD terrain made it easier for the planner to find trajectories to goal.
}
}
\label{fig:spot_ood_qualitative}
\vspace*{-0.2in}
\end{figure}

\begin{figure}[t]
\centering
\includegraphics[width=\linewidth, trim={0cm 0.cm 0cm 0.cm},clip]{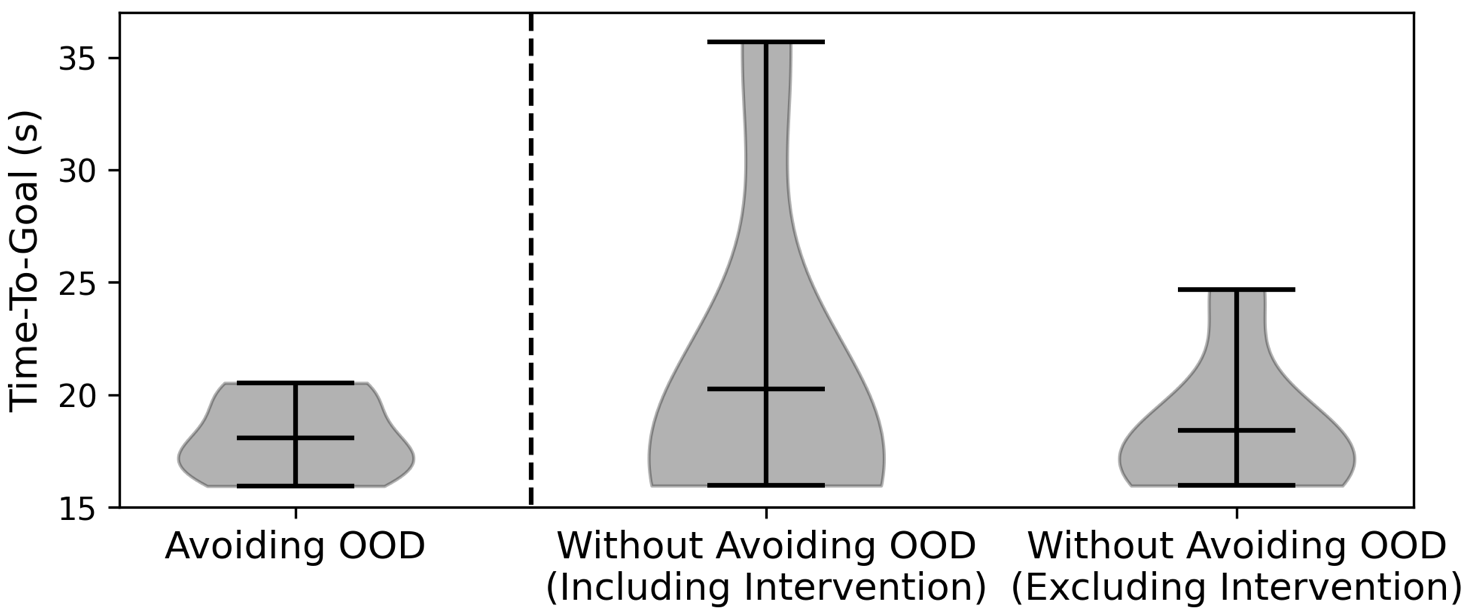}
\caption{
\edit{
Distributions of time-to-goal for the OOD avoidance tests over 6 trials (3 round trips), with maximum, average, and minimum values. By avoiding OOD terrain, the planner was less susceptible to local minima and achieved better time-to-goal by avoiding terrain with features unseen during training.
}
}
\label{fig:spot_ood_quantitative_results}
\vspace*{-0.2in}
\end{figure}

\edit{
Compared to the indoor setting, the outdoor experiments introduce more diverse terrain types and uncertainty in perception due to lighting changes and rough motions. In addition to benchmarking the planners' ability to handle aleatoric uncertainty, the outdoor tests also demonstrate the benefits of mitigating epistemic uncertainty by avoiding OOD terrain, as well as the applicability of our approach on a legged robot. 
}

\subsubsection{\edit{Experiment Setup}}
An overview of the outdoor setup is shown in Fig.~\ref{fig:spot_exp_setup_highlevel}. A Boston Dynamics Spot robot was fitted with a RealSense D455, an Ouster OS0 lidar, and an Nvidia Jetson AGX Orin with good power efficiency but less powerful computation than the computers used in previous experiments. 
The unicycle model~\eqref{eq:unicycle} was used for this experiment, and traction values were obtained by comparing the commanded velocities and Spot's built-in odometry. 
\edit{The environment model was built using a semantic octomap~\cite{asgharivaskasi2021active} that fused lidar points and segmented RGB images based on the 24 semantic categories in the RUGD dataset~\cite{RUGD2019IROS}.}
The traction model was trained based on 5 minutes of walking data with the proposed loss function~\eqref{eq:uce_uemd2_h} with the same weights used for the indoor experiment. The learned traction distributions are selectively visualized in Fig.~\ref{fig:spot_exp_setup_highlevel}a to highlight multi-modality.
As shown in Fig.~\ref{fig:spot_exp_setup_highlevel}b, we chose 2 start-goal pairs for testing the planners and assessing the benefits of avoiding OOD terrain, respectively. All planners avoided the terrain with elevation greater than 1.4~m via auxiliary penalties, and the baseline assigned soft costs for the grass and bush semantic types  with elevations less than 1.4~m. While the 1.4~m height threshold is much higher than the robot's step height, the selected test environments did not have short and rigid obstacles in order to analyze the planners' ability to handle tall vegetation. The robot maintained a semantic octomap with 0.2~m resolution by fusing semantic pointclouds generated from projecting semantic images to lidar pointcloud for accurate depth. Due to limited GPU resources shared by semantic classification, traction prediction, and motion planning, the planners could only reliably plan at 5~Hz with 8~s look-ahead and 800 control rollouts, and \cvarcost{} was only allowed 200 traction map samples. The maximum linear and angular velocities were 1~m/s and 90~degree/s.

\subsubsection{\edit{Aleatoric Uncertainty Results}}
\edit{
The qualitative and quantitative results comparing planners' abilities to mitigate the risk due to aleatoric uncertainty are shown in Fig.~\ref{fig:spot_planner_qualitative} and  Fig.~\ref{fig:spot_planner_quantitative}. We considered 3 round trips to and from the goal (6 trials in total) for each method.
}
Overall, \cvardyn{} with $\alpha=0.9$ achieved the best time-to-goal and success rate, consistent with the indoor experiments in Sec.~\ref{sec:results:indoor_exp_rc}. The \cvarcost{} planner was more conservative by staying far from the bushes. In comparison, the baseline and WayFAST both suffered from noisy real-world traction, causing wide turns. Notably, when the soft penalty for grass and bush semantic types was too high, the baseline planner was stuck in local minima, thus requiring human interventions and long mission time.

\subsubsection{\edit{Epistemic Uncertainty Results}}
\edit{
Different from previous experiments, the goal of the OOD terrain avoidance experiment is to show the benefit of mitigating the risk due to epistemic uncertainty. Therefore, we only used the proposed planner \cvardyn{} with $\alpha=0.9$, but similar conclusions still hold if we change the underlying local planner to \cvarcost{} or another baseline method to mitigate the risk due to aleatoric uncertainty. We executed 3 round trips in total.
}

The qualitative and quantitative results for the OOD avoidance experiments are shown in Fig.~\ref{fig:spot_ood_qualitative} and Fig.~\ref{fig:spot_ood_quantitative_results}.
We considered the terrain as OOD if the normalized densities for the traction predictor's latent features fell below 0 \edit{(i.e., the 0-th percentile of the densities observed for all the training data), but a more conservative threshold may be used based on empirical tuning.} 
\edit{
Compared to the training environment shown in Fig.~\ref{fig:spot_exp_setup_highlevel}, the test environment shown in Fig.~\ref{fig:spot_ood_qualitative} contained much taller vegetation for which we did not collect training data for. As a result, the traction predictions for the tall vegetation terrain produced high epistemic uncertainty and the associated terrain was marked as OOD.
}
Without avoiding OOD terrain, the robot was more prone to getting stuck in local minima and required human interventions to drive to the robot to areas with feasible trajectories to goal. In contrast, the planner that avoided OOD terrain achieved better time-to-goal without requiring human interventions.

\subsection{\edit{Takeaways From the Hardware Experiments}}\label{sec:hw_exp_takeaways}
\edit{
In summary, the hardware experiments have demonstrated that the proposed \cvardyn{} is an attractive choice in practice, without incurring extra computation required by \cvarcost{} that samples additional traction maps or requiring human expertise in designing semantics-based costs for potentially a large variety of terrain types. In addition, the ability to estimate epistemic uncertainty allows us to identify and avoid OOD terrain with unreliable traction predictions, thus improving navigation success rate and reducing human interventions. 
}

\section{\edit{Limitations \& Future Work}}
\edit{
From the modeling standpoint, this work focused on 2D robot models, but models with six degrees of freedom are needed for more challenging terrain~\cite{lee2023AleatoricEpistemicKinodynamicModel, datar2023learning, sharma2023ramp}. In addition, we used a semantic octomap~\cite{asgharivaskasi2021active} to model the environment, but computationally cheaper alternatives~\cite{erni2023mem, ewen2022probfriction} can be used instead. 
Moreover, our work relies on the accuracy of the semantic segmentation module, so the proposed pipeline may fail if the test environments look too different from the training environments (e.g., due to lighting and seasonal changes). Therefore, risk due to the uncertainty in the perception modules need to be addressed separately~\cite{ancha2024icra}.}

\edit{
From the data collection standpoint, this work required empirical traction distributions for training, which may be difficult to attain for high-dimensional features such as RGB images. While the proposed loss can be used to train against instantaneous traction measurements directly, the performance benefits of using EMD$^2$-based loss need to be reassessed. Moreover, uncertainty-guided data collection methods~\cite{kim2023bridging, endo2022active}) can be used to collect informative training samples.
}

\edit{
From the planning standpoint, this work proposed to simulate state trajectories using the CVaR of traction, but more investigations are needed to generalize the idea to systems with more parameters and different performance metrics. Moreover, our planner avoids OOD terrain in new environments, but online adaptation can be performed~\cite{Frey-RSS-23} if human supervision is available. Lastly, the proposed approach can be paired with a global planner that exploits far-field knowledge~\cite{chen2023learning}.
}

\section{Conclusion}
\edit{This work proposed EVORA, a unified framework for uncertainty-aware traversability learning based on evidential deep learning and risk-aware planning based on CVaR. EVORA models uncertain terrain traction via empirical distributions (aleatoric uncertainty) and identifies OOD terrain based on densities of traction predictor's latent features (epistemic uncertainty).} By leveraging the proposed uncertainty-aware squared Earth Mover's Distance loss, we improved the network's prediction accuracy, OOD detection performance, and the downstream navigation performance. To handle aleatoric uncertainty, \edit{the proposed risk-aware planner simulates} state trajectories based on the left-tail CVaR of the traction distributions. To handle epistemic uncertainty, we proposed to assign auxiliary costs to terrain whose latent features have low densities, leading to higher navigation success rates. The overall pipeline was analyzed via extensive simulations and hardware experiments, demonstrating improved navigation performance across different ground robotic platforms.

\appendices
\section{UCE Loss and Dirichlet Entropy~\cite{natpn}}\label{appendix:uce_details}
\edit{
Given $q=\dir(\bm{\beta})$ and the target PMF $\mathbf{y}$:
\begin{align}
 L^\text{UCE}(q, \mathbf{y}) \defeq&~\E_{\mathbf{p}\sim q} \Bigg[  - \sum_{b=1}^B y_b~\log p_b \Bigg] \label{eq:uce_weighted_sum_of_p} \\
    =& - \sum_{b=1}^B y_b (\digammafun(\beta_b) - \digammafun(\beta_0))
\end{align}
where $\digammafun$ is the digamma function and $\beta_0\defeq \sum_{b=1}^B \beta_b$ is the overall evidence.
In addition, the entropy of $q$ is:
\begin{equation}\label{eq:dir_h}
    H(q) = \log \mathcal{B} (\bm{\beta}) + (\beta_0 - B) \digammafun (\beta_0) - \sum_{b=1}^B (\beta_b-1)\digammafun(\beta_b)
\end{equation}
where $\mathcal{B}$ denotes the beta function.
}

\section{Proof of Theorem 1}\label{appendix:uemd2_proof}
We proceed directly from the definition of $\uemdsq$~\eqref{eq:uemd2_def} and simplify the notation by making the expectation over $\mathbf{p}\sim\dir(\bm{\beta})$ implicit. Recall that $\mathbf{y}$ is the target PMF, $\cumsum(\cdot)$ is the cumulative sum operator, and we denote $\cumsum_b(\cdot)$ as the $b$-th entry of the cumulative sum vector.
\begin{align}
&\uemdsq(\bm{\beta}, \mathbf{y}) 
\defeq \E \big[ \emdsq (\mathbf{p}, \mathbf{y}) \big] \\
&= \E \bigg[ \sum_{b=1}^B \big( \cumsum_b(\mathbf{p}) - \cumsum_b(\mathbf{y}) \big)^2 \bigg]\\
&= \sum_{b=1}^B \E \bigg( \sum_{i=1}^b p_i - \sum_{i=1}^b y_i \bigg)^2\\
&= \sum_{b=1}^B  \bigg[ \E \bigg( \sum_{i=1}^b p_i \bigg)^2 - 2\,\E \sum_{i=1}^b p_i\underbrace{\sum_{i=1}^b y_i}_{\textstyle\small\cumsum_b(\mathbf{y})}~+~\E \underbrace{\bigg( \sum_{i=1}^b y_i \bigg)^2}_{\textstyle\small\cumsum_b(\mathbf{y})^2} \bigg].
\end{align}
After separating out the constant additive term $\cumsum_b(\mathbf{y})^2$, expanding the remaining terms, and moving the expectation inside the summation, we obtain:
\begin{align}
=&\sum_{b=1}^B \bigg(
    \sum_{i=1}^b \underbrace{\E[p_i^2]}_{\red{\text{(a)}}}~+~2 \hspace{-1em}\sum_{1\leq i<j\leq b} \underbrace{\E[p_i p_j]}_{\red{\text{(b)}}}
    -~2~\cumsum_b(\mathbf{y}) \sum_{i=1}^b \underbrace{\E[p_i]}_{\red{\text{(c)}}} \bigg)\nonumber\\
&+\cumsum(\mathbf{y})\tr\cumsum(\mathbf{y}).\label{eq:uemd:init_expansion}
\end{align}
The terms \red{(a-c)} in \eqref{eq:uemd:init_expansion} can be easily derived in closed-form based on the standard properties of a Dirichlet distribution (namely, the variance, covariance and mean):
\begin{align}
\red{\text{(a) }} \E[p_i^2]&= \variance(p_i) + \E[p_i]^2 \\
&=\frac{\beta_i (\beta_0-\beta_i)}{\beta_0^2(\beta_0+1)} + \frac{\beta_i^2}{\beta_0^2} \\
&= \frac{\beta_i+\beta_i^2}{\beta_0(\beta_0+1)} \label{eq:uemd:a}
\end{align}
\edit{where $\beta_0\defeq \sum_{b=1}^B \beta_b$}. When $i\not=j$,
\begin{align}
\red{\text{(b) }} \E[p_i p_j] &= \covariance (p_i, p_j) + \E[p_i] \E[p_j] \\
&= \frac{-\beta_i\beta_j}{\beta_0^2(\beta_0 + 1 )} + \frac{\beta_i \beta_j}{\beta_0^2}\\
&= \frac{\beta_i\beta_j}{\beta_0(\beta_0 + 1)}.\label{eq:uemd:b}
\end{align}
Lastly, $\red{\text{(c) }}\E[p_i] = \frac{\beta_i}{\beta_0}$. 
Substituting \red{(a-c)} in~\eqref{eq:uemd:init_expansion}, we obtain
\begin{align}
=& \sum_{b=1}^B \bigg(
    \sum_{i=1}^b \frac{\beta_i+\beta_i^2 }{\beta_0(\beta_0+1)} + 2 \hspace{-0.5em}\sum_{1\leq i<j\leq b} \frac{\beta_i\beta_j}{\beta_0(\beta_0 + 1)} \nonumber\\
    &\quad -2~\cumsum_b(\mathbf{y}) \sum_{i=1}^b \frac{\beta_i}{\beta_0} \bigg) + \cumsum(\mathbf{y})\tr\cumsum(\mathbf{y}) \\
=&~ \frac{1}{\beta_0(\beta_0+1)}\sum_{b=1}^B \bigg(
    \sum_{i=1}^b \beta_i + \sum_{i=1}^b \beta_i^2 + 2\sum_{1\leq i<j\leq b} \beta_i\beta_j  \bigg )\nonumber\\
    &   - \frac{2}{\beta_0} \sum_{b=1}^B \bigg( \cumsum_b(\mathbf{y}) \sum_{i=1}^b \beta_i \bigg) + \cumsum(\mathbf{y})\tr\cumsum(\mathbf{y}) \\
=&~ \frac{1}{\beta_0(\beta_0+1)}\sum_{b=1}^B \bigg(
    \underbrace{\sum_{i=1}^b \beta_i}_{\textstyle\small\cumsum_b(\bm{\beta})} + \underbrace{\Big(\sum_{i=1}^b \beta_i\Big)^2}_{\textstyle\small\cumsum_b(\bm{\beta})^2} \bigg )\nonumber\\
    &   - \frac{2}{\beta_0} \sum_{b=1}^B \bigg( \cumsum_b(\mathbf{y}) \underbrace{\sum_{i=1}^b \beta_i}_{\textstyle\small\cumsum_b(\bm{\beta})} \bigg) +~ \cumsum(\mathbf{y})\tr\cumsum(\mathbf{y}) \\
=&~ \frac{\cumsum(\bm{\beta})}{\beta_0}\tr \frac{ \cumsum(\bm{\beta}) + \mathbb{1}_B}{(\beta_0+1)} 
 - 2\frac{\cumsum(\bm{\beta})}{\beta_0}\tr  \cumsum(\mathbf{y}) + \cumsum(\mathbf{y})\tr\cumsum(\mathbf{y}) \\
=&~ \cumsum(\overline{\mathbf{p}})\tr \frac{\cumsum(\bm{\beta})+\mathbb{1}_B}{\beta_0+1} + \eta(q, \mathbf{y})
\end{align}
where $\overline{\mathbf{p}} = \bm{\beta} / \beta_0 = E_{\mathbf{p}\sim \dir(\bm{\beta})}[\mathbf{p}]$ and $\eta$ is defined in~\eqref{eq:emd_l_term}.
\\
\noindent$\square$

\bibliographystyle{IEEEtran}
\bibliography{bibs}

\end{document}